\documentclass[10pt,twocolumn,letterpaper]{article}

\usepackage{cvpr}              

\usepackage{multirow}

\usepackage{framed}
\usepackage{siunitx}
\usepackage{latexsym}
\usepackage{colortbl}
\usepackage[dvipsnames]{xcolor}
\usepackage{nicefrac}
\usepackage{booktabs}
\usepackage{fnpct}
\usepackage{amsfonts}
\usepackage[T1]{fontenc}
\usepackage{bold-extra}
\usepackage{amsmath}
\usepackage{amssymb}
\usepackage{bm}
\usepackage{graphicx}
\usepackage{mathtools}
\usepackage{microtype}
\usepackage{multirow}
\usepackage{multicol}
\usepackage{xpatch}
\usepackage{tcolorbox}
\usepackage{latexsym,comment}
\usepackage[normalem]{ulem}

\newcommand{\ours}{MASS}
\newcommand{\oursdata}{MASS-Bench}

\usepackage{tikz}
\usepackage{multicol}
\usepackage{graphicx}
\usepackage{xcolor}

\def\savelastnode{\pgfextra\edef\tmpA{\tikzlastnode}\endpgfextra}
\def\restorelastnode{\pgfextra\edef\tikzlastnode{\tmpA}\endpgfextra}

\tikzstyle{mybox} = [draw=black, fill=yellow!20, thick,
    rectangle, rounded corners, inner sep=10pt, inner ysep=20pt]
\tikzstyle{mynewbox} = [draw=black, fill=yellow!20, thick,
    rectangle, rounded corners, inner sep=5pt, inner ysep=3pt]
\tikzstyle{fancytitle} =[fill=black, text=white]
\tikzstyle{title} = [append after command={%
    \savelastnode node[fancytitle,right=10pt] at (\tikzlastnode.north west)%
    {\large{#1}}\restorelastnode}]

\newcommand{\bluebold}[1]{\textcolor{Blue}{\large{\textbf{#1}}}}

\usepackage{soul}

\usepackage[accsupp]{axessibility}

\definecolor{cvprblue}{rgb}{0.21,0.49,0.74}
\usepackage[pagebackref,breaklinks,colorlinks,allcolors=cvprblue]{hyperref}

\def\paperID{9689} 
\def\confName{CVPR}
\def\confYear{2026}

\title{\ours{}: Motion-Aware Spatial–temporal Grounding for Physics Reasoning and Comprehension in Vision-Language Models}

\author{
Xiyang Wu$^{1,2}$ \quad
Zongxia Li$^{1}$ \quad
Jihui Jin$^{2}$ \quad
Guangyao Shi$^{3}$ \quad
Gouthaman KV$^{2}$ \quad
Vishnu Raj$^{2}$ \\
Nilotpal Sinha$^{2}$ \quad
Jingxi Chen$^{1}$ \quad
Fan Du$^{2}$ \quad
Dinesh Manocha$^{1}$ \\[4pt]
$^{1}$University of Maryland \quad
$^{2}$Dolby Laboratories \quad
$^{3}$University of Southern California
}

\newcommand{\wxy}[1]{\textbf{\textcolor{blue}{$\leftarrow$ #1}}}

\begin{document}
\maketitle
\begin{abstract}

Vision Language Models (VLMs) perform well on standard video tasks but struggle with physics-related reasoning that involves motion dynamics and spatial interactions. 
%
%
We present a novel approach to address this gap by translating physical-world context cues into interpretable representations aligned with VLMs’ perception, comprehension, and reasoning. 
We present \ours{}, a model-agnostic approach that injects spatial–temporal signals into the VLM language space via depth-based 3D encoding and visual grounding, coupled with a motion tracker for object dynamics. 
We also contribute a comprehensive benchmark, \oursdata{}, consisting of $4,350$ real-world and AIGC videos and $8,361$ free-form video question–answering pairs focused on physics-related comprehension tasks, with detailed annotations including visual detections and grounding over sub-segments, as well as full-sequence 3D motion tracking of entities.
To strengthen cross-modal alignment and reasoning, we apply reinforcement fine-tuning to \ours{}. 
Experiments and ablations show that our refined VLMs outperform comparable and baselines of larger size, and prior state-of-the-art (SoTA) models, achieving performance comparable to close-source SoTA VLMs  ($2\%$ difference) like Gemini-2.5-Flash on physics reasoning and comprehension.

\end{abstract}    
\section{Introduction}


Vision Language Models (VLMs) demonstrate strong reasoning and comprehension in standard video tasks such as captioning~\cite{li2024wolf}, event recognition~\cite{li2025eventvl} and scene understanding~\cite{yu2025evaluating}. However, they struggle with complex visual cues that involve intertwined 3D spatial layouts~\cite{cai2025holisticevaluationmultimodalllms, yang2025thinking}, motion patterns~\cite{chen2024motionllm}, and temporal dynamics~\cite{zhou2025vlm4d, fu2025video}. 
Achieving robust physical understanding requires VLMs not only to perceive visual cues, but also to internalize real-world physical principles and commonsense expectations about object behavior~\cite{chow2025physbench}. Such physical reasoning, central to human-level video understanding, remains challenging, as models must connect visual evidence with underlying physical dynamics and reason about whether observed events align with or violate real-world physics.

Unlike tasks such as segmentation or object grounding, where the supervision is explicit and localized, real-world physics must be inferred from indirect and often ambiguous visual evidence. 
%
%
For example, an apple falling and a person standing still are both influenced by gravity but differ substantially in spatial configuration, motion patterns, and temporal structure. 
%
%
As a result, VLMs often fail to generalize across diverse physical processes~\cite{balazadeh2025physics, li2025videohallu}, a challenge further magnified by the scarcity of datasets with dense spatiotemporal and motion-level annotations~\cite{deng2025motion, liang2025fine}. 
Without explicit supervision, models tend to memorize superficial correlations rather than develop physics-grounded reasoning~\cite{chi2025chimera}. 
Effective physics comprehension requires several demanding prerequisites, spatial and temporal understanding, motion tracking, object detection, and visual grounding, yet these components are rarely annotated in sufficient detail~\cite{li2025sti, wang2025spatialvid}. 

\begin{figure*}
    \centering
    \includegraphics[width=0.95\linewidth]{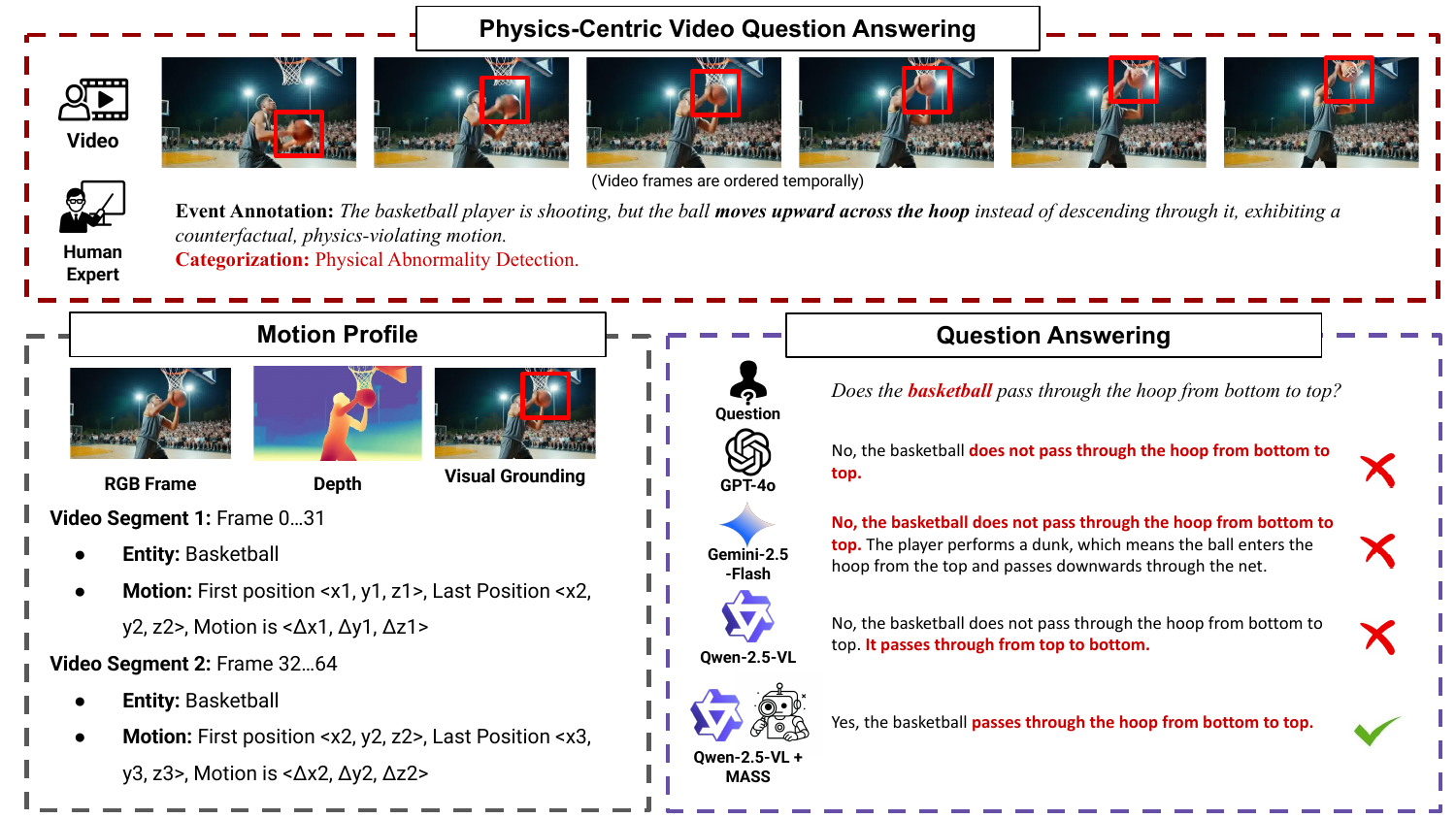}
    \vspace{-5pt}
\caption{\textbf{Physics-Centric Video Question Answering.}
Physics-aware video comprehension is challenging, as VLMs must capture fine-grained spatial–temporal cues and integrate them for higher-level reasoning. \ours{} introduces a motion-aware spatial–temporal grounding module that explicitly encodes object motions and scene dynamics into the language space. By enriching VLMs with structured spatial, temporal, and semantic signals, \ours{} significantly improves downstream reasoning, including motion and action understanding, physical-process inference, and abnormality detection (\textit{e.g.}, identifying the counterfactual upward motion of a basketball). \ours{} outperforms strong SoTA models such as GPT-4o and Gemini-2.5-Flash, demonstrating robust physics comprehension and reasoning across diverse tasks.}
    \label{fig:teaser}
    \vspace{-12pt}
\end{figure*}


Most VLMs are trained primarily on real-world videos, where object motion usually follows physical laws. 
Although such data helps models capture common motion patterns, it also reinforces strong visual and language priors: models often assume that observed events are physically plausible because they match frequent patterns in the training data. 
This becomes more problematic with AIGC videos, where implausible trajectories, inconsistent depth, and temporally incoherent motion are common. 
Prior studies show that VLMs often miss or hallucinate such physics violations~\cite{bansal2025videophy, li2025videohallu}. For example, a model may infer that an apple is falling simply because it appears near a tree, relying on prior expectations rather than visual evidence.

These challenges highlight the need for going beyond raw video pixels and high-level textual supervision. Instead of training VLMs from scratch, which is infeasible and very costly in time and computational resource, one more data-efficient way is to augment VLMs with structured spatial and temporal representations that capture fine-grained object motions, interactions, and geometry. Recent advances in motion grounding~\cite{deng2025motion}, visual grounding~\cite{li2026mmzeroselfevolvingmultimodelvision,li2025self}, detailed temporal modeling~\cite{ye2025re}, and action analysis~\cite{du2025motionsight, hong2025motionbench} demonstrate that smaller specialized models can provide highly accurate spatial–temporal cues. 
However, existing approaches rarely explore how such modules can be systematically integrated into VLMs to elevate higher-level physics reasoning. Meanwhile, prior physics-oriented datasets~\cite{li2025videohallu, li2025worldmodelbench, bansal2025videophy} primarily rely on coarse annotations, limiting their capacity to induce deep, mechanism-level understanding.



\noindent
\textbf{Main Results:}
Our approach, \ours{}, aims to bridge the gap between the structured dynamics of the physical world and VLMs’ perception, comprehension, and reasoning abilities. We leverage expert models’ spatial and motion representations to explicitly ground entities and encode their 3D trajectories, enabling VLMs to reason about motion, interactions, and spatial constraints rather than relying on coarse priors. To evaluate these capabilities, we design a comprehensive free-form video QA benchmark covering physics phenomena with both factual and critical-thinking questions. For post-training, we explore both supervised fine-tuning (SFT) and reinforcement fine-tuning (RFT) using Group Relative Policy Optimization (GRPO) to strengthen cross-modal alignment and reasoning under complex physical contexts. The novel contributions of our work include:

\begin{itemize}
    \item We propose a benchmark, \oursdata{}, of $4,350$ real and AIGC videos with $8,361$ free-form QA pairs on spatial-temporal comprehension, physics-related reasoning, and abnormality detection. It includes entity-level motion grounded annotation with visual grounding over spatial and temporal dimension and dense spatial-motion representations across videos.
    \item We introduce \ours{}, a model-agnostic, motion-aware spatial–temporal grounding algorithm for physics reasoning and comprehension. \ours{} explicitly grounds and represents spatial–motion information of entities in video, integrating visual grounding with structured spatial and temporal representations through a spatiotemporal awareness module. This enables VLMs to capture and encode object dynamics that are otherwise inaccessible from raw prompt inputs.
    \item We post-train VLMs using GRPO to improve their comprehension and reasoning and cross-modality alignment over physics-related video phenomena. Experiments and ablations show that our refined VLMs outperform comparable and larger baselines, and prior state-of-the-art models, by $8.7\%$ and $6.0\%$, achieving performance close to closed-source SoTA VLMs like Gemini-2.5-Flash and validating the effectiveness of our approach.
\end{itemize}


\if
Abnormalities in AI-generated content (AIGC) videos~\cite{li2025worldmodelbench, qin2024worldsimbench, li2025videohallu} remain a deeply rooted and challenging problem. Human prompts used to generate AIGC contexts are typically brief and high-level, while the resulting videos are rich in both static and dynamic details. This mismatch introduces a significant gap between input and output in text-to-video (T2V) tasks. Consequently, AIGC videos often suffer from fundamental limitations, including misalignment with instructions, motion inconsistency for dynamic entities, and higher-level violations of physics and common sense.


Traditionally, video generation evaluation has relied on metric-based methods such as Inception Score (IS)~\cite{barratt2018note} and Fréchet Inception Distance (FID)~\cite{heusel2017gans}. These approaches, along with early AIGC video evaluation efforts like VideoScore~\cite{he2024videoscore}, primarily focus on assessing visual quality and realism rather than measuring the faithfulness of generated content to instructions or real-world plausibility. 
While realism remains an essential criterion, state-of-the-art models~\cite{brooks2024video, yang2024cogvideox, veo3} already produce high-fidelity videos but often fail in physics and commonsense consistency.


Moreover, score- and metric-based methods lack explainability: their models or rule-based computations offer no transparent rubric for how scores are derived. Although the numbers may seem intuitive as proxies for video quality, their ambiguity hinders error attribution and correction. High-level video comprehension, such as reasoning about physics or commonsense, requires models to provide explanations and reasoning chains~\cite{wei2022chain}, not just output a single score. This lack of interpretability limits the diagnosis and improvement of AIGC videos and further complicates training, as models struggle to detect and resolve issues.


The limitations of metric-based evaluation and persistent issues in video quality and faithfulness highlight the need for more advanced methods with stronger comprehension and reasoning for AIGC videos. A natural solution is VLM-as-a-judge~\cite{chen2024VLM}, where VLMs assess videos through rigorous question answering, probing abnormalities in physics or commonsense using evaluation rubrics or predefined templates. This leverages their large-scale knowledge and reasoning abilities. Prior works~\cite{li2025worldmodelbench, duan2025worldscore, bansal2025videophy} explored prompt design or fine-tuning with human-annotated datasets, showing promise for this pipeline. However, these early efforts lack deeper insight into physics and commonsense, which require visual grounding on queried points and robust spatial–motion understanding of the 4D physical world.
\wxy{}


The current use of AIGC video datasets with physics and commonsense artifacts only superficially leverages VLMs for evaluation and fails to adequately address hallucination in video understanding. Unlike real-world video tasks, AIGC video QA requires counterfactual reasoning that often conflicts with real-world knowledge. Moreover, flawed AIGC videos demand precise question constraints, yet instruction-following VLMs may hallucinate by relying on their parameterized knowledge rather than actual visual inputs, leading to bias toward observed context. \wxy{is the observed content the current input or previous training data thats free of physics violations? I think this point is important to stress} Such instruction-following hallucinations, combined with weak spatial and motion comprehension, exacerbate the challenge of rigorous evaluation of physics and commonsense artifacts in AIGC videos.

This work aims to bridge the gap between AIGC video understanding and real-world knowledge, enhancing VLMs’ ability to evaluate AIGC videos and detect physics and commonsense violations with stronger spatial, motion, and reasoning capabilities to reduce hallucinations. We address this gap by integrating richer training datasets that capture physical and commonsense comprehension across both real and AIGC videos, framed as video QA tasks probing abnormalities. Our approach incorporates model-agnostic spatial and motion awareness modules into existing VLM frameworks, and applies supervised and reinforcement fine-tuning to improve cross-modal alignment and reasoning over complex contexts.
Our main contributions include:
\fi
\section{Related Work}

\textbf{Physical Reasoning and Abnormality Detection in Videos:}
Despite rapid progress in AIGC video generation, generated videos still often contain prompt mismatches and violations of physics or commonsense, from earlier models such as LaVIE~\cite{wang2025lavie}, SORA~\cite{brooks2024video}, and CogVideoX~\cite{yang2024cogvideox} to more recent systems including VEO3~\cite{veo3}, Wan2.2~\cite{wan2025}, and COSMOS~\cite{agarwal2025cosmos}. This has motivated growing work on abnormality evaluation and mitigation. VideoScore~\cite{he2024videoscore} trains synthetic evaluators aligned with human judgment, while recent benchmarks~\cite{li2025worldmodelbench, qin2024worldsimbench, motamed2025generative, chow2025physbench, li2025videohallu, zhang2025morpheus} support systematic evaluation of physics and commonsense reasoning. WorldScore~\cite{duan2025worldscore} provides a unified protocol over controllability, quality, and dynamics, and CRAVE~\cite{sun2025content} studies content-rich assessment through text--temporal fusion and motion-fidelity modeling. On the mitigation side, prior work has explored physics-based cues~\cite{kang2024far}, chain-of-thought reasoning in PhyT2V~\cite{xue2025phyt2v}, action-centric evaluation and correction in VideoPhy-2~\cite{bansal2025videophy}, and motion-aware plausibility judgment in TRAVL~\cite{motamed2025travl}. Still, reliably identifying and addressing physical and commonsense failures remains an open challenge.

\noindent
\textbf{Spatial, Temporal, and Motion Understanding in VLMs:}
Recent work has improved video understanding in VLMs along spatial, temporal, and motion dimensions. For spatial reasoning, SpatialRGPT~\cite{cheng2024spatialrgpt}, VG-LLM~\cite{zheng2025learning}, and LayoutVLM~\cite{sun2025layoutvlm} strengthen 3D perception through depth cues, video-grounded geometry, and layout optimization. For temporal modeling, SlowFast-LLaVA~\cite{xu2024slowfast} captures both local detail and long-range dynamics, TVS~\cite{fan2025video} improves alignment with temporally relevant segments, and T*~\cite{ye2025re} studies long-video retrieval via spatial search. Other efforts focus on motion and action understanding, including OpenMixer~\cite{bao2025exploiting}, MotionSight~\cite{du2025motionsight}, and GroundMoRe~\cite{deng2025motion}. However, physics-centric reasoning remains underexplored, as it requires jointly integrating spatial, temporal, and motion cues with physical commonsense.

\noindent
\textbf{Video Understanding in VLMs:}
Recent VLMs, including Qwen-VL~\cite{bai2025qwen2} and InternVL~\cite{zhu2025internvl3}, have improved general video understanding, yet remain limited on complex spatial, temporal, and long-video reasoning~\cite{li2025survey}. LongVLM~\cite{weng2024longvlm} improves long-video modeling through hierarchical token merging, while VideoMind~\cite{liu2025videomind} uses a role-based agent with Chain-of-LoRA for efficient temporal reasoning. Video-Holmes~\cite{cheng2025video} highlights persistent failures on multi-segment reasoning, and EgoLife~\cite{yang2025egolife} supports long-context retrieval for egocentric video assistants. Other work addresses hallucination and reasoning, including MASH-VLM~\cite{bae2025mash}, MVoT~\cite{li2025imagine}, and large-scale 3D spatial VQA data in~\cite{chen2024spatialvlm}. Reinforcement-based training has also shown promise, as in Video-R1~\cite{feng2025video} and VideoChat-R1~\cite{li2025videochat}.

\section{\oursdata{}: A Motion-Grounded Physics Reasoning and Comprehension Benchmark}
\label{sec:dataset}

In this section, we introduce the details of our benchmark. High-quality training data is essential for enabling video models to understand the physical world, yet existing resources fall short when tasks extend beyond scene-level captioning to entity-level spatiotemporal reasoning and factual comprehension, such as physics understanding, commonsense reasoning, or abnormality detection in AIGC videos. This gap largely stems from the lack of datasets that (1) provide fine-grained, spatially and temporally consistent annotations of entities across frames, and (2) include a balanced mixture of \underline{\textit{positive}} examples that follow real-world physical dynamics and \underline{\textit{negative}} examples that intentionally violate physical laws. In the absence of such enriched and balanced supervision, VLMs remain limited to surface-level perception; by contrast, datasets meeting these criteria enable more robust, physics-aware reasoning over dynamic video content.

\noindent
\begin{figure*}
    \centering
    \includegraphics[width=0.88\linewidth]{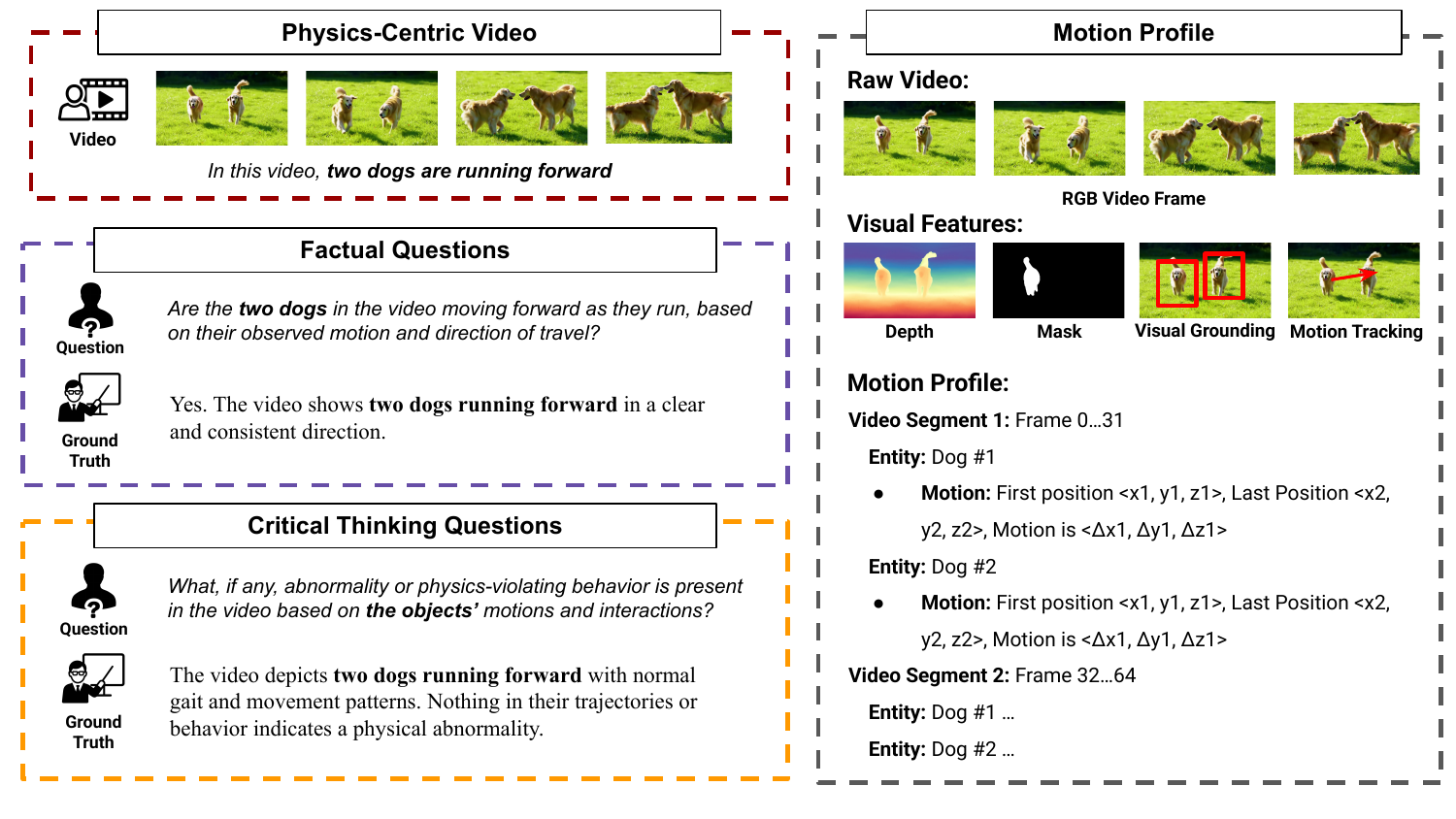}
\vspace{-10pt}
\caption{\textbf{Data Exhibition of \oursdata{}.} \oursdata{} provides two question types, including \textit{factual} and \textit{critical-thinking}, to evaluate physics-driven video understanding. For each video–question–answer pair, we supply rich \textbf{motion-grounding annotations}, including\textit{ temporal segmentation}, \textit{entity-level visual grounding}, \textit{temporal profiles across the full video}, and \textbf{motion attributes} such as \textit{first/last positions} and \textit{3D displacement vectors}. These structured spatial–temporal cues transform complex physics-related perception into interpretable representations that support more reliable physical reasoning. Additional dataset details are provided in the supplementary material.}
    \label{fig:dataset}
    \vspace{-12pt}
\end{figure*}

\noindent
\textbf{Data Collection:} \oursdata{} is a free-form video question-answering dataset designed to strengthen VLMs’ physics understanding, built from real-world videos sourced from Motion-Sight~\cite{du2025motionsight} and ActivityNet~\cite{caba2015activitynet}, alongside AIGC videos collected from VideoPhy2~\cite{bansal2025videophy}, VideoHallu~\cite{li2025videohallu}, and generative models such as COSMOS~\cite{agarwal2025cosmos}, Wan2.2~\cite{wan2025}, and Sora~\cite{brooks2024video}. 
\oursdata{} consists of $4,350$ videos, composed by $8,361$ free-form video question-answering pairs related to physics-related comprehension tasks. The dataset includes both positive and negative demonstrations with detailed human annotations providing balanced coverage of labels and reasoning steps. This design enables VLMs to learn correct physical processes from real-world videos, while also recognizing artifacts and violations that commonly arise in AIGC videos but are rare in natural scenes.


\noindent
\textbf{Questions:} As shown in Figure~\ref{fig:dataset}, our dataset contains two types of reasoning questions: \textit{Factual} questions that assess models’ grounding of real versus hypothetical scenarios, asking whether a physical process follows real-world principles or identifying the underlying mechanism, and, for AIGC videos, determining whether the physics aligns with reality or exhibits artifact-driven abnormalities.
and \textit{Critical-thinking} questions that require inferring causes, intentions, or detecting physical or commonsense abnormalities beyond explicit annotations. We re-annotate multiple-choice and labeled physics tasks into open-ended QA using Claude-4-Sonnet, and all ground-truth answers are further verified by humans to ensure physical consistency and accessibility to both humans and VLMs.








\noindent
\textbf{Categorization:}
To capture the spectrum of physics-related cognition required in \oursdata{}, we group all questions into five categories ordered from basic to advanced: \textit{Spatial Understanding} (SU, 33.3\%), \textit{Temporal Understanding} (TU, 19.5\%), \textit{Motion \& Action Recognition} (MAR, 14.4\%), \textit{Physics Comprehension} (PC, 15.6\%), and \textit{Physical Abnormality Detection} (PA, 17.1\%). VLMs must first establish spatial and temporal awareness, then recognize motion patterns, before progressing to higher-level physics comprehension and real-world violation detection. The full category definitions and distribution are provided in Table 5 of the supplementary material.

\noindent
\textbf{Motion Grounding.}
Beyond the question, video, and ground-truth answer, each pair includes motion-grounding annotations: temporal segmentation indexed by frame ranges, entity-level visual grounding with bounding boxes, temporal profiles tracking each entity across the full video, and per-segment motion attributes (first/last 3D positions and motion vectors). These cues convert physics-intensive perceptual challenges into structured textual representations for more reliable reasoning. Full annotation details are provided in the supplementary material.

\begin{figure*}
    \centering
    \includegraphics[width=0.9\linewidth]{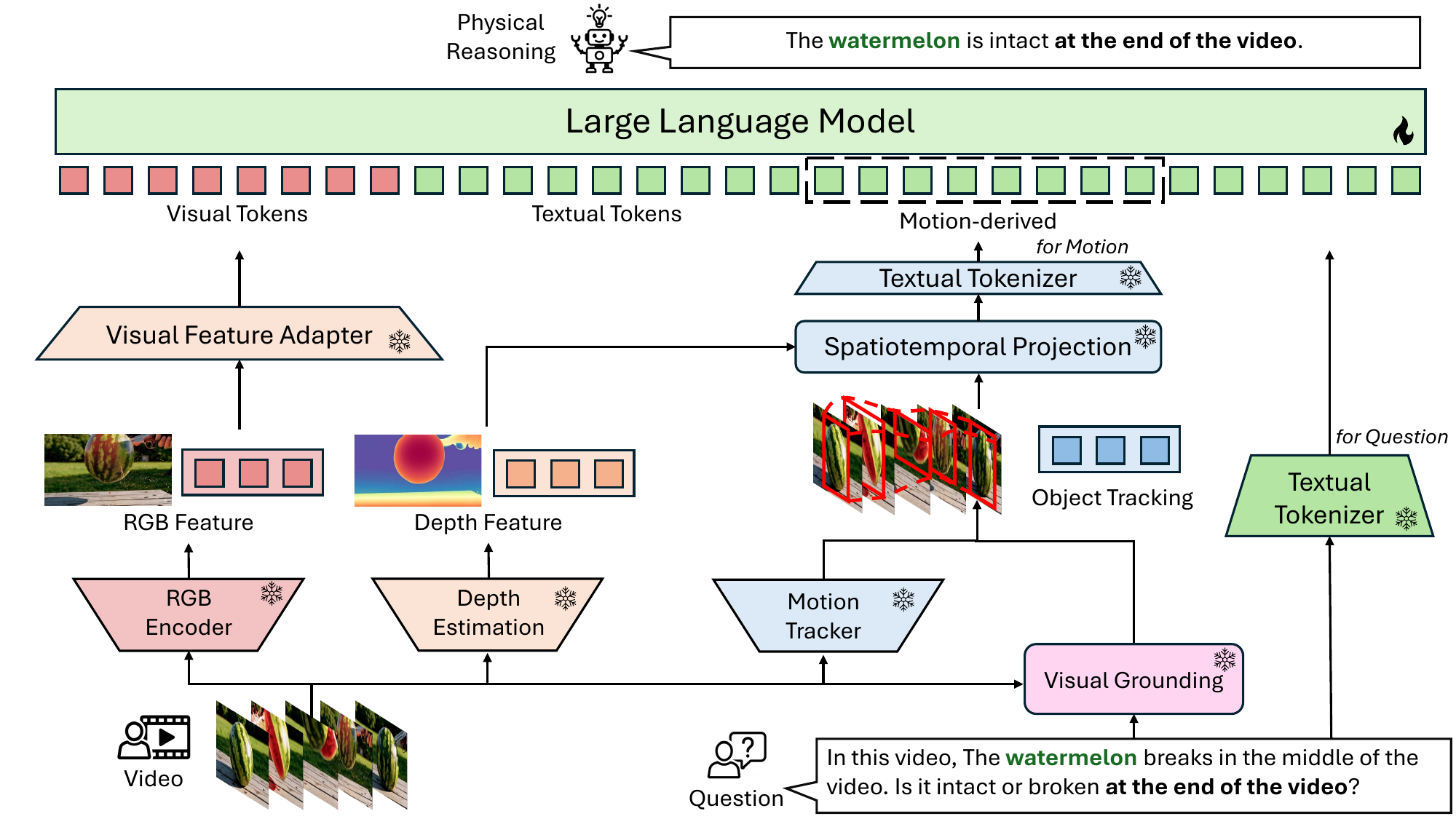}
    \caption{\textbf{Overview of \ours{}:} We use a model-agnostic framework to enhance video understanding with explicit spatial and motion awareness. Beyond standard visual transformer encoders that process video inputs (e.g., LLaVA-OneVision~\cite{li2024llava}, Qwen2.5-VL~\cite{bai2025qwen2}), we introduce a visual grounding module to better align queried entities with corresponding visual cues. Depth estimation captures spatial geometry, while motion tracking captures temporal dynamics across frames. These cues are fused into entity-centric motion traces and converted into \emph{motion-derived textual tokens} that are appended to the textual input stream. During post-training, we keep the spatial--temporal encoders frozen, \textit{i.e.}, used off-the-shelf without parameter updates, and apply reinforcement fine-tuning (RFT) only to improve the LLM backbone's comprehension of the additional multimodal information.}
    \label{fig:pipeline}
    \vspace{-10pt}
\end{figure*}

\section{ \ours{}: Model-Agnostic Approach}
\label{sec:method}


High-quality human annotations are essential for strengthening VLMs’ understanding of physics-related visual contexts, yet data alone cannot address their core limitations in modeling spatial layouts and motion dynamics. Effective physical reasoning requires isolating relevant cues from noisy video content. While one solution is to retrain models with heavy preprocessing pipelines, this incurs substantial cost. Instead, we propose a more data-efficient alternative: augmenting VLMs with dedicated spatial and motion encoders that explicitly extract and represent key visual signals. As shown in Figure~\ref{fig:pipeline}, our model-agnostic design combines lightweight architectural refinement with targeted training to enhance spatiotemporal perception, cross-modal alignment, and physics-aware reasoning, ultimately reducing hallucinations and improving physical comprehension.


\subsection{Entity-Centric Visual Grounding}

Understanding physics-related dynamics in videos often depends on accurately identifying and tracking the relevant entities across time. This remains challenging because videos contain dense spatial--temporal content, while language queries are often underspecified about which objects are relevant. Grounding is therefore critical for localizing queried entities, preserving their motion and spatial context, and filtering out irrelevant information. It also helps mitigate the information loss introduced by LLM-based VLMs such as LLaVA~\cite{liu2023visual}, which compress rich visual signals into a limited set of language-aligned tokens.

We begin by semantically extracting the key components from each probing question to narrow the target for visual grounding. Using Grounding-DINO~\cite{liu2024grounding}, we detect bounding boxes for the queried entities and apply a dynamic temporal-resolution scheme that automatically selects an appropriate sub-sequence length based on the input video’s duration. This segmentation strategy increases temporal coherence within each chunk, improving object tracking and downstream processing while balancing accuracy and efficiency. It also preserves grounding consistency over time and reveals temporal artifacts such as sudden appearance or disappearance, which are common in AIGC videos. For each detected entity, we construct a time-aligned grounding profile. Finally, we apply SAM2~\cite{ravi2024sam, ren2024grounded} to generate segmentation masks for the grounded regions, enabling subsequent spatial–temporal analysis.

\subsection{Spatial Motion Feature Extraction}

Physical reasoning depends on accurate knowledge of where entities are, how they move, and how their interactions evolve over time. After identifying the key entities referenced by the probing questions, the main challenge lies in the VLM’s limited ability to comprehend their spatial positions, appearances, and dynamics across frames. Without explicit spatial–temporal understanding, VLMs struggle to track entities, recognize actions, and maintain frame-level consistency, ultimately hindering higher-level physical reasoning. To address these limitations, we extract explicit spatial–temporal representations for each queried entity by integrating domain-specialized visual encoders capable of translating raw spatial–motion cues into structured features accessible to the VLM.



Leveraging the narrowed-down, visually grounded entities identified by the probing questions, we perform motion tracking for each entity within its video sub-sequences using CoTracker3~\cite{karaev2024cotracker3}, which tracks arbitrary point trajectories inside the grounded regions. For each entity, we compute the spatial deviations across time, including averaged motion vectors and the first/last tracked positions. In parallel, we estimate per-frame depth using Depth Anything V2~\cite{depth_anything_v2} to provide spatial awareness. By aligning sampled RGB frames with their corresponding depth maps, we project the tracked motion vectors over each tracking points throughout the entity into 3D space, yielding both 3D positions and full spatio-temporal motion trajectories for each entity.


\subsection{Visual Feature Representation}

After encoding visual features across different modules, the key challenge lies in integrating heterogeneous spatial–motion outputs into the VLM pipeline without disrupting its native architecture. Prior work~\cite{wu2024autohallusion, li2025videohallu} shows that LLM backbones better comprehend abstract, factual, and physics-related concepts when such information is expressed in the language space rather than as raw visual features.

Motivated by this, we fuse visual cues through natural-language representations aligned with the tokenized questions. We compute these features because VLMs cannot reliably infer fine-grained spatial layouts or motion dynamics from compressed visual embeddings; providing explicit, structured cues supplies the physical information needed for stable and interpretable reasoning. Our visual-feature serialization converts spatial–motion signals into structured motion-trajectory sequences for each entity, including 3D start/end positions, motion vectors, bounding boxes within each video sub-segment, and associated temporal indices. These attributes are expressed through predefined natural-language templates (detailed in the supplementary material), enabling the VLM to access precise spatial–temporal information in the modality it handles most effectively.

\subsection{Training Pipeline}

Although visual grounding and spatial–motion features provide essential cues, VLMs cannot directly leverage these signals for higher-level reasoning because their representations are misaligned with those learned during pre-training. As a result, even precise motion trajectories, spatial positions, and temporal indices such as knowing an object is intact at position A at time 1 and broken at position B at time 2 cannot be meaningfully synthesized without additional alignment. 
The model must bridge this representational gap to integrate structured cues, filter irrelevant details, and reason over scene evolution rather than merely retrieving or describing visual content.


We perform post-training to enhance the VLM’s reasoning capabilities to bridge this gap. Chain-of-thought (CoT) reasoning~\cite{wei2022chain} helps the model process complex temporal and physical concepts more effectively. By encouraging intermediate reasoning steps, CoT aligns the VLM’s decision process with the multi-stage deductions needed for physical comprehension. This training enables the VLM to operate on grounded spatial–motion features, integrate temporal changes, and produce coherent, physics-aware explanations rather than treating cues as isolated descriptors. We evaluate both Supervised Fine-tuning (SFT) and Reinforcement Fine-tuning (RFT). For RFT, we adopt Temporal Group Relative Policy Optimization (T-GRPO)~\cite{feng2025video}, which adds temporal-aware rewards on top of vanilla GRPO~\cite{shao2024deepseekmath}, a key component for video understanding. We further incorporate ROUGE~\cite{lin2004rouge} for semantic coherence, a temporal reward to strengthen temporal reasoning, and a format reward to encourage proper CoT generation.
\section{Empirical Results}
\label{sec:results}
For our experiments, we evaluate 12 widely used state-of-the-art VLMs as baselines (e.g., GPT-4o and Gemini-2.5-Flash), covering 7B--38B parameter scales. We report performance comparisons between these baselines and VLMs augmented with \ours{} in Table~\ref{tab:physics_breakdown}, and present ablations on our backbone models and post-training strategies in Table~\ref{tab:ablation}. For evaluation, we use GPT-4.1-Mini as an automated judge to assess the factual correctness between each model’s answer and the ground-truth annotations regarding these physical comprehension and reasoning questions provided. 
We provide additional ablation studies are provided in the supplementary material, including real-world video QA generalization (Section E), noise-induced robustness analysis (Section F), and inference efficiency measurements (Section G).

\begin{table}[t]
\centering
\footnotesize
\resizebox{\columnwidth}{!}{
\begin{tabular}{l|cccccc}
\toprule
Model & SU & TU & MAR & PC & PA & Overall \\
\midrule
\multicolumn{7}{c}{\textit{Baselines}} \\
\midrule
\texttt{VideoLLaVA}                & 36.39 & 31.76 & 21.01 & 38.55 & 17.76 & 30.72 \\
\texttt{InternVL3.5-8B-Flash}      & 47.22 & 45.25 & 39.40 & \textbf{58.65} & 32.76 & 44.68 \\
\texttt{InternVL3-8B}              & 48.20 & 49.89 & 39.29 & 57.20 & 32.47 & 45.59 \\
\texttt{InternVL3.5-14B-Flash}     & 47.68 & 49.23 & 41.96 & 57.52 & 31.03 & 45.71 \\
\texttt{LLaVA-OneVision-7B}        & 49.18 & 51.99 & 38.12 & 56.02 & 30.46 & 45.79 \\
\texttt{Qwen2.5-VL-7B}             & 55.87 & \textbf{53.03} & \textbf{54.89} & 52.36 & 30.21 & 52.30 \\
\texttt{Qwen3-VL-8B}               & \textbf{57.29} & 50.77 & 51.92 & \textbf{58.65} & \textbf{37.07} & \textbf{53.23} \\
\midrule
\multicolumn{7}{c}{\textit{R1-finetuned Reasoning VLMs}} \\
\midrule
\texttt{VideoChat-R1}             & 51.20 & 48.66 & 47.34 & 52.30 & 31.09 & 47.05 \\
\texttt{Video-R1}                 & \textbf{57.49} & \textbf{55.19} & \textbf{47.65} & \textbf{63.53} & \textbf{33.91} & \textbf{53.08} \\
\midrule
\multicolumn{7}{c}{\textit{Larger / Close-Source}} \\
\midrule
\texttt{InternVL3.5-38B-Flash}     & 52.19 & 49.88 & 43.29 & 61.45 & 32.71 & 48.71 \\
\texttt{GPT-4o}                    & 54.47 & 55.13 & 55.62 & 60.15 & 31.32 & 52.81 \\
\texttt{Qwen3-VL-32B}              & 58.87 & 50.59 & 48.05 & 59.84 & 38.63 & 53.44 \\
\texttt{Gemini-2.5-Flash}          & \textbf{65.82} & \textbf{59.65} & \textbf{68.57} & \textbf{65.15} & \textbf{43.80} & \textbf{63.18} \\
\midrule
\multicolumn{7}{c}{\textit{VLMs with \ours{}}} \\
\midrule
\texttt{LLaVA-OneVision-7B + \ours{}} & 61.26 & \textbf{65.41} & 58.15 & 61.45 & 42.37 & 59.24 \\
\texttt{Qwen2.5-VL-7B + \ours{}}      & \textbf{63.52} & 63.06 & \textbf{64.07} & \textbf{65.06} & \textbf{45.79} & \textbf{61.93} \\
\bottomrule
\end{tabular}
}
\vspace{-5pt}
\caption{\textbf{Accuracy (\%) across categories in \oursdata{}}: Performance of VLMs across five physics-related categories—Spatial Understanding (SU), Temporal Understanding (TU), Motion \& Action Recognition (MAR), Physics Comprehension (PC), and Physical Abnormality Detection (PA). Models are grouped by family and sorted by overall accuracy. Scaling to 32B–38B parameters yields only marginal gains over 8B counterparts, whereas integrating \ours{} with 7B models substantially improves performance and achieves accuracy comparable to Gemini-2.5-Flash by leveraging explicit spatial–temporal representations.}

\label{tab:physics_breakdown}
\vspace{-10pt}
\end{table}

\subsection{Baseline Insights: Why Motion-Aware Spatial-temporal Grounding Matters}

\textbf{Spatial--Temporal Signals Are Critical for Physics Understanding.}
We evaluate baseline VLMs across a wide parameter range, from 7B--8B models in the LLaVA~\cite{liu2023visual}, Qwen-VL~\cite{bai2025qwen2}, and InternVL~\cite{zhu2025internvl3} families, up to 32B--38B variants and close-source models such as GPT-4o and Gemini-2.5-Flash. Under the same evaluation setting, neither newer variants nor larger models consistently perform better. For instance, Qwen2.5-VL-7B and Qwen3-VL-8B achieve similar results, and InternVL3.5-8B-Flash shows no clear advantage over InternVL3-8B. This trend also appears at larger scales: InternVL3.5-38B-Flash improves by only $4.0\%$ over its 8B version, while Qwen3-VL-32B remains comparable to Qwen3-VL-8B. By contrast, incorporating \ours{} yields substantial improvements. Qwen2.5-VL-7B+\ours{} and LLaVA-OneVision-7B+\ours{} exceed the previous best model, Qwen3-VL-8B, by $8.7\%$ and $6.0\%$, respectively. Qwen2.5-VL-7B+\ours{} also matches Gemini-2.5-Flash overall and outperforms it on physics abnormality detection, the most reasoning-heavy category. Together, these results show that explicit spatial--temporal representations supply physical cues that standard VLMs often miss, and are therefore key to reliable physics reasoning.


\noindent
\textbf{Training Data Quality is Crucial for Reasoning.}
We also include two R1-finetuned VLMs, Video-R1 and VideoChat-R1, in Table~\ref{tab:physics_breakdown}, both trained with GRPO to enhance reasoning. However, neither model shows clear improvement on physics comprehension tasks compared with non–GRPO fine-tuned VLMs. In contrast, integrating \ours{} with Qwen2.5-VL-7B and LLaVA-OneVision-7B yields substantially higher performance, highlighting the importance of high-quality training data. Detailed annotations, along with both positive (successful) and negative (failure) demonstrations, are essential for strengthening reasoning and counteracting language priors when tackling out-of-domain physics tasks. Without such data, RFT alone offers limited gains and may even reinforce hallucinations arising from misinterpreting visual evidence.

\subsection{Ablation Study: The Role of Reasoning}
\label{exp:ablation}

\textbf{Reasoning Dominates Beyond the Base Model.} Table~\ref{tab:ablation} compares post-training results on two backbone VLMs, Qwen2.5-VL-7B~\cite{bai2025qwen2} and LLaVA-OneVision-7B~\cite{li2024llava}. Across both models, enhanced reasoning contributes far more than the backbone choice itself: pairing Qwen2.5-VL-7B with \ours{} yields a $9.6\%$ improvement, while LLaVA-OneVision-7B combined with \ours{} achieves a larger $13.5\%$ gain. After post-training, the base model’s influence becomes secondary, as physics comprehension, especially challenging tasks like physics abnormality detection, demands integrated reasoning over visual physics cues, spatial–temporal structure, and commonsense priors. This synergy, rather than raw model capacity, ultimately drives the performance improvements.

\noindent
\textbf{Visual Evidence and Motion Context Are Complementary.} We ablate the input modality using three settings: (i) \textbf{Text-only} (motion profile + question), (ii) \textbf{Visual-only} (video tokens + question), and (iii) \textbf{Both} (video tokens + motion profile + question, our default). Table~\ref{tab:input_variants_sft_grpo} reports these results alongside the SFT vs.\ GRPO comparison. Text-only consistently underperforms visual-only across all models (e.g., GPT-5.2: 42.09\% vs.\ 56.17\%; Gemini-2.5-Pro: 43.39\% vs.\ 63.41\%), confirming that many questions are inherently video-dependent. While the motion profile provides useful temporal summaries, it cannot replace the spatial detail and grounded evidence in the original video. Consistent with this, about 26\% of text-only outputs are judged as \textit{uncertain} by the LLM judge. 
Using both modalities yields the best results. Under the Both setting, GRPO + \ours{} improves Qwen2.5-VL-7B by 9.6\% and LLaVA-OneVision-7B by 13.0\%. This shows that the motion profile is most effective as complementary context rather than a substitute for video. The benefit also depends on model capacity: large models such as GPT-5.2 and Gemini-2.5-Pro can exploit motion-profile text even without post-training, whereas 7B models gain little from the added text unless they are further trained to integrate it. This pattern suggests that the improvement does not come from extra textual context alone, but from learning to align motion-profile cues with visual evidence for better physics reasoning.

\noindent
\textbf{Why RFT Matters: SFT Alone Is Not Enough.} Table~\ref{tab:ablation} compares SFT and RFT strategies. 
SFT yields limited benefit overall: it slightly reduces performance for Qwen2.5-VL-7B by 0.3\% and brings only a modest 4.2\% gain for LLaVA-OneVision-7B, leaving both close to the off-the-shelf models. By contrast, GRPO produces large and consistent gains across both backbones, reaching 61.93\% on Qwen2.5-VL-7B and 59.24\% on LLaVA-OneVision-7B. We believe this is because GRPO optimizes task-level rewards directly, making it better suited than SFT for learning physical reasoning that depends on temporal evidence and causal consistency. The gains are especially clear on SU, TU, MAR, and PC, where stronger reasoning is needed to connect motion cues with physical dynamics. PA is the main exception, where SFT performs better in our reruns, likely because PA depends more on answer calibration: the model must not only detect the abnormality, but also describe it in a precise free-form way that matches the reference. Even so, GRPO remains the stronger strategy overall.

\begin{table}[t]
\centering
\footnotesize
\resizebox{\columnwidth}{!}{
\begin{tabular}{l|ccc|cc}
\toprule
 & \multicolumn{3}{c|}{\textbf{Input Variants}} & \multicolumn{2}{c}{\textbf{Training Strategy}} \\
\multirow{-2}{*}{\textbf{Model}} & \textbf{Text} & \textbf{Visual} & \textbf{Both} & \textbf{SFT} & \textbf{GRPO} \\
\midrule
\texttt{GPT-5.2}                     & 42.09 & 56.17 & 59.46 & --    & --    \\
\texttt{Gemini-2.5-Pro}              & 43.39 & 63.41 & 64.30 & --    & --    \\
\midrule
\texttt{Qwen2.5-VL-7B}               & 32.14 & 52.30 & 52.38 & 50.40 & 52.89 \\
\quad + \texttt{GRPO + \ours{}}       & 39.24 & 52.94 & 61.93 & 52.01 & 61.93 \\
\midrule
\texttt{LLaVA-OneVision-7B}          & 31.69 & 45.79 & 46.21 & 47.60 & 50.33 \\
\quad + \texttt{GRPO + \ours{}}       & 35.63 & 46.72 & 59.24 & 50.03 & 59.24 \\
\bottomrule
\end{tabular}
}
\vspace{-5pt}
\caption{\textbf{Input modality ablation (left) and training strategy comparison (right) for models with and without \ours{}.} Results are reported as accuracy (\%) on \oursdata{}.}
\label{tab:input_variants_sft_grpo}
\vspace{-5pt}
\end{table}


\begin{table}[t]
\centering
\footnotesize
\resizebox{\columnwidth}{!}{
\begin{tabular}{l|cccccc}
\toprule
Model & SU & TU & MAR & PC & PA & Overall \\
\midrule
\texttt{Qwen2.5-VL-7B}             & 55.87 & 53.03 & 54.89 & 52.36 & 30.21 & 52.30 \\
\quad + \texttt{SFT}               & 53.68 & 54.32 & 48.10 & 59.02 & \textbf{47.54} & 52.01 \\
\quad + \texttt{GRPO}              & \textbf{63.52} & \textbf{63.06} & \textbf{64.07} & \textbf{65.06} & 45.79 & \textbf{61.93} \\
\midrule
\texttt{LLaVA-OneVision-7B}        & 49.18 & 51.99 & 38.12 & 56.02 & 30.46 & 45.79 \\
\quad + \texttt{SFT}               & 47.81 & 51.22 & 49.84 & 50.00 & \textbf{49.26} & 50.03 \\
\quad + \texttt{GRPO}              & \textbf{61.26} & \textbf{65.41} & \textbf{58.15} & \textbf{61.45} & 42.37 & \textbf{59.24} \\
\bottomrule
\end{tabular}
}
\vspace{-5pt}
\caption{\textbf{Ablation of post-training strategies.} Accuracy (\%) of base VLMs, SFT variants, and GRPO-enhanced models across five physics-related categories. GRPO provides substantial gains over both base models and SFT in most categories.}
\label{tab:ablation}
\vspace{-10pt}
\end{table}

\noindent
\textbf{Training Data and Module Ablations.} 
We ablate the GRPO training data and the motion representation under the same training budget. Table~\ref{tab:grpo_data_module_ablation} shows two main takeaways. First, the balanced \emph{Both} setting works best on both backbones. \emph{Real-only} training brings almost no gain over the off-the-shelf models, while \emph{Syn-only} helps more but still remains below \emph{Both}. This suggests that the two data sources help in different ways: real videos add natural variation in appearance and motion, while synthetic data more clearly expose the physical patterns the model needs to learn. Second, 3D motion matters. Replacing the full 3D motion representation with \emph{2D motion} lowers accuracy for both models. Because this ablation only removes the depth-aware projection, the drop is not just about having fewer useful tokens; it shows that depth-aware motion provides information that 2D motion alone misses. Overall, the gains come from combining complementary training data with motion cues that better reflect the physical structure of the scene.

\begin{table}[t]
\centering
\footnotesize
\resizebox{\columnwidth}{!}{
\begin{tabular}{l|cc|cc|c}
\toprule
 & \multicolumn{2}{c|}{\textbf{Baselines (\ours{})}} & \multicolumn{2}{c|}{\textbf{Training Data}} & \multicolumn{1}{c}{\textbf{Modules}} \\
\multirow{-2}{*}{Model (+ \ours{})} & w/ & w/o  & Real-only & Syn-only & 2D Motion \\
\midrule
\texttt{Qwen2.5-VL-7B} & 61.93 & 52.30 & 52.40 & 56.30 & 56.50 \\
\texttt{LLaVA-OneVision-7B}    & 59.24 & 45.79 & 49.92 & 53.41 & 52.30 \\
\bottomrule
\end{tabular}
}

\vspace{-5pt}
\caption{\textbf{Ablation of GRPO training data and depth module.} Accuracy (\%) on \oursdata{}. \textit{w/} and \textit{w/o} denote the full \ours{} pipeline (real+syn data, 3D motion) and the base model without \ours{}, respectively. \textit{Real-only} and \textit{Syn-only} vary the training data, while \textit{2D Motion} removes the depth-aware 3D projection.}
\label{tab:grpo_data_module_ablation}
\vspace{-10pt}
\end{table}

\subsection{Discussion}
\label{exp:discussion}

\textbf{The Impact of Motion-aware Spatiotemporal Grounding.}
In Table~\ref{tab:physics_breakdown} and Table~\ref{tab:ablation}, we observe that across model architectures, scales, and post-training strategies, VLMs equipped with our motion-aware spatiotemporal grounding outperform their baselines and reach performance comparable to Gemini-2.5-Flash. These gains stem from explicitly integrating spatiotemporal signals, as standard VLM encoders struggle to utilize or correlate spatial and temporal cues effectively. Providing structured representations of key video dynamics enables stronger physics comprehension and reasoning while filtering irrelevant or distracting visual information.


\noindent
\textbf{Reasoning is Essential for Understanding Physics.}
Motion-aware spatiotemporal grounding is necessary but not sufficient for physics comprehension. Reasoning is the key mechanism that integrates and interprets the grounded information. As shown in Table~\ref{tab:ablation}, comparing SFT and GRPO reveals that even with motion-aware spatiotemporal signals, learning purely from input–output correlations (as in SFT) does not help—and often degrades—performance on physics reasoning tasks. Given the complexity and interdependence of physical processes, VLMs cannot grasp the underlying dynamics through grounding alone and may lose focus during question answering. In contrast, reasoning via chain-of-thought encourages VLMs to reflect on grounded cues and combine them with commonsense knowledge, effectively bridging the gap between visual perception and real-world physical understanding.


\noindent
\textbf{Challenges in Physical Abnormality Detection Remain.}
A key observation from Table~\ref{tab:physics_breakdown} and Table~\ref{tab:ablation} is that Qwen2.5-VL-7B and LLaVA-OneVision-7B integrated with \ours{} outperform all other VLMs, including Gemini-2.5-Flash, on physical abnormality detection. However, compared with more grounded physics comprehension tasks, abnormality detection is inherently more difficult. It requires VLMs to overcome hallucinations rooted in their learned priors about real-world physics~\cite{li2025videohallu} and to reason about why a visual process violates physical laws. While our method yields a notable $12\%$ improvement on this task, indicating the promise of motion-aware spatiotemporal grounding for enhancing perception and reasoning, further advances are needed to robustly handle these challenging cases.

\section{Conclusion, Limitations and Future Work}

We present a model-agnostic approach that bridges the gap between raw video content and physics-aware reasoning by translating physical-world cues into structured representations compatible with VLM inference. We introduce \oursdata{}, to our knowledge the first benchmark pairing positive and negative video demonstrations with QA on spatial-temporal comprehension, physics-based reasoning, and abnormality detection, supported by fine-grained visual detections, sub-segment grounding, and 3D motion tracking. Our method, \ours{}, injects spatial–temporal signals into the VLM language space via depth-based 3D encoding, visual grounding, and motion-aware trajectory modeling, yielding consistent gains over strong baselines and recent SoTA VLMs on both real-world and AIGC videos.

\noindent\textbf{Limitations.}
Although post-training improves reasoning, it does not fully solve physical understanding in real-world settings. Performance may still decline in crowded scenes, where multi-object tracking becomes more demanding, and hallucinations persist in reasoning-heavy cases when subtle visual evidence is misread or overshadowed by language priors. \ours{}-Bench is entity-centric, focusing on the top-$K$ targets, and thus does not evaluate highly crowded scenes. Robust handling of strong camera ego-motion also remains open. Encouragingly, \ours{} remains reasonably stable under noisy motion cues, with gradual rather than catastrophic degradation (see Section F in Supplementary Material).


\noindent\textbf{Future Work.}
Future work will improve long-range motion grounding and temporal reasoning in longer videos, and make multi-object tracking more scalable in crowded scenes through better spatial--temporal profiling. We also plan to extend the benchmark to denser multi-entity settings and stronger camera motion. Finally, expanding the training data with more diverse positive and negative examples may further reduce hallucinations and improve coverage of more challenging physics understanding tasks.

{
    \small
    \bibliographystyle{ieeenat_fullname}
    \bibliography{main}

@article{feng2025video,
  title={Video-r1: Reinforcing video reasoning in mllms},
  author={Feng, Kaituo and Gong, Kaixiong and Li, Bohao and Guo, Zonghao and Wang, Yibing and Peng, Tianshuo and Wu, Junfei and Zhang, Xiaoying and Wang, Benyou and Yue, Xiangyu},
  journal={arXiv preprint arXiv:2503.21776},
  year={2025}
}

@article{li2025videochat,
  title={Videochat-r1: Enhancing spatio-temporal perception via reinforcement fine-tuning},
  author={Li, Xinhao and Yan, Ziang and Meng, Desen and Dong, Lu and Zeng, Xiangyu and He, Yinan and Wang, Yali and Qiao, Yu and Wang, Yi and Wang, Limin},
  journal={arXiv preprint arXiv:2504.06958},
  year={2025}
}

@article{li2024llava,
  title={Llava-onevision: Easy visual task transfer},
  author={Li, Bo and Zhang, Yuanhan and Guo, Dong and Zhang, Renrui and Li, Feng and Zhang, Hao and Zhang, Kaichen and Zhang, Peiyuan and Li, Yanwei and Liu, Ziwei and others},
  journal={arXiv preprint arXiv:2408.03326},
  year={2024}
}

@article{li2025videohallu,
  title={VideoHallu: Evaluating and Mitigating Multi-modal Hallucinations on Synthetic Video Understanding},
  author={Li, Zongxia and Wu, Xiyang and Shi, Guangyao and Qin, Yubin and Du, Hongyang and Zhou, Tianyi and Manocha, Dinesh and Boyd-Graber, Jordan Lee},
  journal={arXiv preprint arXiv:2505.01481},
  year={2025}
}

@article{li2025worldmodelbench,
  title={Worldmodelbench: Judging video generation models as world models},
  author={Li, Dacheng and Fang, Yunhao and Chen, Yukang and Yang, Shuo and Cao, Shiyi and Wong, Justin and Luo, Michael and Wang, Xiaolong and Yin, Hongxu and Gonzalez, Joseph E and others},
  journal={arXiv preprint arXiv:2502.20694},
  year={2025}
}

@misc{ren2024grounded,
      title={Grounded SAM: Assembling Open-World Models for Diverse Visual Tasks}, 
      author={Tianhe Ren and Shilong Liu and Ailing Zeng and Jing Lin and Kunchang Li and He Cao and Jiayu Chen and Xinyu Huang and Yukang Chen and Feng Yan and Zhaoyang Zeng and Hao Zhang and Feng Li and Jie Yang and Hongyang Li and Qing Jiang and Lei Zhang},
      year={2024},
      eprint={2401.14159},
      archivePrefix={arXiv},
      primaryClass={cs.CV}
}

@article{bansal2025videophy,
  title={Videophy-2: A challenging action-centric physical commonsense evaluation in video generation},
  author={Bansal, Hritik and Peng, Clark and Bitton, Yonatan and Goldenberg, Roman and Grover, Aditya and Chang, Kai-Wei},
  journal={arXiv preprint arXiv:2503.06800},
  year={2025}
}

@article{kang2024far,
  title={How far is video generation from world model: A physical law perspective},
  author={Kang, Bingyi and Yue, Yang and Lu, Rui and Lin, Zhijie and Zhao, Yang and Wang, Kaixin and Huang, Gao and Feng, Jiashi},
  journal={arXiv preprint arXiv:2411.02385},
  year={2024}
}

@article{barratt2018note,
  title={A note on the inception score},
  author={Barratt, Shane and Sharma, Rishi},
  journal={arXiv preprint arXiv:1801.01973},
  year={2018}
}

@article{heusel2017gans,
  title={Gans trained by a two time-scale update rule converge to a local nash equilibrium},
  author={Heusel, Martin and Ramsauer, Hubert and Unterthiner, Thomas and Nessler, Bernhard and Hochreiter, Sepp},
  journal={Advances in neural information processing systems},
  volume={30},
  year={2017}
}

@article{duan2025worldscore,
  title={Worldscore: A unified evaluation benchmark for world generation},
  author={Duan, Haoyi and Yu, Hong-Xing and Chen, Sirui and Fei-Fei, Li and Wu, Jiajun},
  journal={arXiv preprint arXiv:2504.00983},
  year={2025}
}

@article{he2024videoscore,
  title={Videoscore: Building automatic metrics to simulate fine-grained human feedback for video generation},
  author={He, Xuan and Jiang, Dongfu and Zhang, Ge and Ku, Max and Soni, Achint and Siu, Sherman and Chen, Haonan and Chandra, Abhranil and Jiang, Ziyan and Arulraj, Aaran and others},
  journal={arXiv preprint arXiv:2406.15252},
  year={2024}
}

@article{qin2024worldsimbench,
  title={Worldsimbench: Towards video generation models as world simulators},
  author={Qin, Yiran and Shi, Zhelun and Yu, Jiwen and Wang, Xijun and Zhou, Enshen and Li, Lijun and Yin, Zhenfei and Liu, Xihui and Sheng, Lu and Shao, Jing and others},
  journal={arXiv preprint arXiv:2410.18072},
  year={2024}
}

@article{wei2022chain,
  title={Chain-of-thought prompting elicits reasoning in large language models},
  author={Wei, Jason and Wang, Xuezhi and Schuurmans, Dale and Bosma, Maarten and Xia, Fei and Chi, Ed and Le, Quoc V and Zhou, Denny and others},
  journal={Advances in neural information processing systems},
  volume={35},
  pages={24824--24837},
  year={2022}
}

@article{brooks2024video,
  title={Video generation models as world simulators},
  author={Brooks, Tim and Peebles, Bill and Holmes, Connor and DePue, Will and Guo, Yufei and Jing, Li and Schnurr, David and Taylor, Joe and Luhman, Troy and Luhman, Eric and others},
  journal={OpenAI Blog},
  volume={1},
  number={8},
  pages={1},
  year={2024}
}

@article{yang2024cogvideox,
  title={Cogvideox: Text-to-video diffusion models with an expert transformer},
  author={Yang, Zhuoyi and Teng, Jiayan and Zheng, Wendi and Ding, Ming and Huang, Shiyu and Xu, Jiazheng and Yang, Yuanming and Hong, Wenyi and Zhang, Xiaohan and Feng, Guanyu and others},
  journal={arXiv preprint arXiv:2408.06072},
  year={2024}
}

@article{veo3,
  title={Veo 3},
  author={Veo-Team},
  url={https://deepmind.google/models/veo/},
 journal={DeepMind Blog},
  year={2025}
}

@article{depth_anything_v2,
  title={Depth Anything V2},
  author={Yang, Lihe and Kang, Bingyi and Huang, Zilong and Zhao, Zhen and Xu, Xiaogang and Feng, Jiashi and Zhao, Hengshuang},
  journal={arXiv:2406.09414},
  year={2024}
}

@article{wan2025,
      title={Wan: Open and Advanced Large-Scale Video Generative Models}, 
      author={Team Wan},
      journal = {arXiv preprint arXiv:2503.20314},
      year={2025}
}

@article{agarwal2025cosmos,
  title={Cosmos world foundation model platform for physical ai},
  author={Agarwal, Niket and Ali, Arslan and Bala, Maciej and Balaji, Yogesh and Barker, Erik and Cai, Tiffany and Chattopadhyay, Prithvijit and Chen, Yongxin and Cui, Yin and Ding, Yifan and others},
  journal={arXiv preprint arXiv:2501.03575},
  year={2025}
}

@article{wang2025lavie,
  title={Lavie: High-quality video generation with cascaded latent diffusion models},
  author={Wang, Yaohui and Chen, Xinyuan and Ma, Xin and Zhou, Shangchen and Huang, Ziqi and Wang, Yi and Yang, Ceyuan and He, Yinan and Yu, Jiashuo and Yang, Peiqing and others},
  journal={International Journal of Computer Vision},
  volume={133},
  number={5},
  pages={3059--3078},
  year={2025},
  publisher={Springer}
}

@article{sun2025content,
  title={Content-rich aigc video quality assessment via intricate text alignment and motion-aware consistency},
  author={Sun, Shangkun and Liang, Xiaoyu and Qu, Bowen and Gao, Wei},
  journal={arXiv preprint arXiv:2502.04076},
  year={2025}
}

@article{motamed2025generative,
  title={Do generative video models understand physical principles?},
  author={Motamed, Saman and Culp, Laura and Swersky, Kevin and Jaini, Priyank and Geirhos, Robert},
  journal={arXiv preprint arXiv:2501.09038},
  year={2025}
}

@inproceedings{xue2025phyt2v,
  title={Phyt2v: Llm-guided iterative self-refinement for physics-grounded text-to-video generation},
  author={Xue, Qiyao and Yin, Xiangyu and Yang, Boyuan and Gao, Wei},
  booktitle={Proceedings of the Computer Vision and Pattern Recognition Conference},
  pages={18826--18836},
  year={2025}
}

@article{wu2024autohallusion,
  title={Autohallusion: Automatic generation of hallucination benchmarks for vision-language models},
  author={Wu, Xiyang and Guan, Tianrui and Li, Dianqi and Huang, Shuaiyi and Liu, Xiaoyu and Wang, Xijun and Xian, Ruiqi and Shrivastava, Abhinav and Huang, Furong and Boyd-Graber, Jordan Lee and others},
  journal={arXiv preprint arXiv:2406.10900},
  year={2024}
}

@article{bai2025qwen2,
  title={Qwen2. 5-vl technical report},
  author={Bai, Shuai and Chen, Keqin and Liu, Xuejing and Wang, Jialin and Ge, Wenbin and Song, Sibo and Dang, Kai and Wang, Peng and Wang, Shijie and Tang, Jun and others},
  journal={arXiv preprint arXiv:2502.13923},
  year={2025}
}

@article{zhu2025internvl3,
  title={Internvl3: Exploring advanced training and test-time recipes for open-source multimodal models},
  author={Zhu, Jinguo and Wang, Weiyun and Chen, Zhe and Liu, Zhaoyang and Ye, Shenglong and Gu, Lixin and Tian, Hao and Duan, Yuchen and Su, Weijie and Shao, Jie and others},
  journal={arXiv preprint arXiv:2504.10479},
  year={2025}
}

@article{liu2025videomind,
  title={VideoMind: A Chain-of-LoRA Agent for Long Video Reasoning},
  author={Liu, Ye and Lin, Kevin Qinghong and Chen, Chang Wen and Shou, Mike Zheng},
  journal={arXiv preprint arXiv:2503.13444},
  year={2025}
}

@article{cheng2025video,
  title={Video-Holmes: Can MLLM Think Like Holmes for Complex Video Reasoning?},
  author={Cheng, Junhao and Ge, Yuying and Wang, Teng and Ge, Yixiao and Liao, Jing and Shan, Ying},
  journal={arXiv preprint arXiv:2505.21374},
  year={2025}
}

@inproceedings{weng2024longvlm,
  title={Longvlm: Efficient long video understanding via large language models},
  author={Weng, Yuetian and Han, Mingfei and He, Haoyu and Chang, Xiaojun and Zhuang, Bohan},
  booktitle={European Conference on Computer Vision},
  pages={453--470},
  year={2024},
  organization={Springer}
}

@inproceedings{bae2025mash,
  title={MASH-VLM: Mitigating Action-Scene Hallucination in Video-LLMs through Disentangled Spatial-Temporal Representations},
  author={Bae, Kyungho and Kim, Jinhyung and Lee, Sihaeng and Lee, Soonyoung and Lee, Gunhee and Choi, Jinwoo},
  booktitle={Proceedings of the Computer Vision and Pattern Recognition Conference},
  pages={13744--13753},
  year={2025}
}

@article{li2025imagine,
  title={Imagine while reasoning in space: Multimodal visualization-of-thought},
  author={Li, Chengzu and Wu, Wenshan and Zhang, Huanyu and Xia, Yan and Mao, Shaoguang and Dong, Li and Vuli{\'c}, Ivan and Wei, Furu},
  journal={arXiv preprint arXiv:2501.07542},
  year={2025}
}

@inproceedings{chen2024spatialvlm,
  title={Spatialvlm: Endowing vision-language models with spatial reasoning capabilities},
  author={Chen, Boyuan and Xu, Zhuo and Kirmani, Sean and Ichter, Brain and Sadigh, Dorsa and Guibas, Leonidas and Xia, Fei},
  booktitle={Proceedings of the IEEE/CVF Conference on Computer Vision and Pattern Recognition},
  pages={14455--14465},
  year={2024}
}

@inproceedings{yang2025egolife,
  title={Egolife: Towards egocentric life assistant},
  author={Yang, Jingkang and Liu, Shuai and Guo, Hongming and Dong, Yuhao and Zhang, Xiamengwei and Zhang, Sicheng and Wang, Pengyun and Zhou, Zitang and Xie, Binzhu and Wang, Ziyue and others},
  booktitle={Proceedings of the Computer Vision and Pattern Recognition Conference},
  pages={28885--28900},
  year={2025}
}

@article{li2026mmzeroselfevolvingmultimodelvision,
  title={MM-Zero: Self-Evolving Multi-Model Vision Language Models From Zero Data},
  author={Li, Zongxia and Du, Hongyang and Huang, Chengsong and Wu, Xiyang and Yu, Lantao and He, Yicheng and Xie, Jing and Wu, Xiaomin and Liu, Zhichao and Zhang, Jiarui and others},
  journal={arXiv preprint arXiv:2603.09206},
  year={2026}
}

@article{li2025self,
  title={Self-rewarding vision-language model via reasoning decomposition},
  author={Li, Zongxia and Yu, Wenhao and Huang, Chengsong and Liu, Rui and Liang, Zhenwen and Liu, Fuxiao and Che, Jingxi and Yu, Dian and Boyd-Graber, Jordan and Mi, Haitao and others},
  journal={arXiv preprint arXiv:2508.19652},
  year={2025}
}

@article{chow2025physbench,
  title={Physbench: Benchmarking and enhancing vision-language models for physical world understanding},
  author={Chow, Wei and Mao, Jiageng and Li, Boyi and Seita, Daniel and Guizilini, Vitor and Wang, Yue},
  journal={arXiv preprint arXiv:2501.16411},
  year={2025}
}

@inproceedings{hong2025motionbench,
  title={Motionbench: Benchmarking and improving fine-grained video motion understanding for vision language models},
  author={Hong, Wenyi and Cheng, Yean and Yang, Zhuoyi and Wang, Weihan and Wang, Lefan and Gu, Xiaotao and Huang, Shiyu and Dong, Yuxiao and Tang, Jie},
  booktitle={Proceedings of the Computer Vision and Pattern Recognition Conference},
  pages={8450--8460},
  year={2025}
}

@article{karaev2024cotracker3,
  title={Cotracker3: Simpler and better point tracking by pseudo-labelling real videos},
  author={Karaev, Nikita and Makarov, Iurii and Wang, Jianyuan and Neverova, Natalia and Vedaldi, Andrea and Rupprecht, Christian},
  journal={arXiv preprint arXiv:2410.11831},
  year={2024}
}

@article{liu2023visual,
  title={Visual instruction tuning},
  author={Liu, Haotian and Li, Chunyuan and Wu, Qingyang and Lee, Yong Jae},
  journal={Advances in neural information processing systems},
  volume={36},
  pages={34892--34916},
  year={2023}
}

@article{zheng2025learning,
  title={Learning from Videos for 3D World: Enhancing MLLMs with 3D Vision Geometry Priors},
  author={Zheng, Duo and Huang, Shijia and Li, Yanyang and Wang, Liwei},
  journal={arXiv preprint arXiv:2505.24625},
  year={2025}
}

@article{shao2024deepseekmath,
  title={Deepseekmath: Pushing the limits of mathematical reasoning in open language models, 2024},
  author={Shao, Zhihong and Wang, Peiyi and Zhu, Qihao and Xu, Runxin and Song, Junxiao and Bi, Xiao and Zhang, Haowei and Zhang, Mingchuan and Li, YK and Wu, Y and others},
  journal={URL https://arxiv. org/abs/2402.03300},
  volume={2},
  number={3},
  pages={5},
  year={2024}
}

@article{du2025motionsight,
  title={MotionSight: Boosting Fine-Grained Motion Understanding in Multimodal LLMs},
  author={Du, Yipeng and Fan, Tiehan and Nan, Kepan and Xie, Rui and Zhou, Penghao and Li, Xiang and Yang, Jian and Yang, Zhenheng and Tai, Ying},
  journal={arXiv preprint arXiv:2506.01674},
  year={2025}
}

@inproceedings{yu2025evaluating,
  title={Evaluating Multimodal Large Language Models on Video Captioning via Monte Carlo Tree Search},
  author={Yu, Linhao and Ji, Xingguang and Liu, Yahui and Kong, Fanheng and Sun, Chenxi and Zhang, Jingyuan and Zhang, Hongzhi and Zhang, Fuzheng and Xiong, Deyi and others},
  booktitle={Proceedings of the 63rd Annual Meeting of the Association for Computational Linguistics (Volume 1: Long Papers)},
  pages={6435--6462},
  year={2025}
}

@inproceedings{deng2025motion,
  title={Motion-grounded video reasoning: Understanding and perceiving motion at pixel level},
  author={Deng, Andong and Chen, Tongjia and Yu, Shoubin and Yang, Taojiannan and Spencer, Lincoln and Tian, Yapeng and Mian, Ajmal Saeed and Bansal, Mohit and Chen, Chen},
  booktitle={Proceedings of the Computer Vision and Pattern Recognition Conference},
  pages={8625--8636},
  year={2025}
}

@inproceedings{ye2025re,
  title={Re-thinking temporal search for long-form video understanding},
  author={Ye, Jinhui and Wang, Zihan and Sun, Haosen and Chandrasegaran, Keshigeyan and Durante, Zane and Eyzaguirre, Cristobal and Bisk, Yonatan and Niebles, Juan Carlos and Adeli, Ehsan and Fei-Fei, Li and others},
  booktitle={Proceedings of the Computer Vision and Pattern Recognition Conference},
  pages={8579--8591},
  year={2025}
}

@inproceedings{liu2024grounding,
  title={Grounding dino: Marrying dino with grounded pre-training for open-set object detection},
  author={Liu, Shilong and Zeng, Zhaoyang and Ren, Tianhe and Li, Feng and Zhang, Hao and Yang, Jie and Jiang, Qing and Li, Chunyuan and Yang, Jianwei and Su, Hang and others},
  booktitle={European conference on computer vision},
  pages={38--55},
  year={2024},
  organization={Springer}
}

@article{ravi2024sam,
  title={Sam 2: Segment anything in images and videos},
  author={Ravi, Nikhila and Gabeur, Valentin and Hu, Yuan-Ting and Hu, Ronghang and Ryali, Chaitanya and Ma, Tengyu and Khedr, Haitham and R{\"a}dle, Roman and Rolland, Chloe and Gustafson, Laura and others},
  journal={arXiv preprint arXiv:2408.00714},
  year={2024}
}

@inproceedings{lin2004rouge,
  title={Rouge: A package for automatic evaluation of summaries},
  author={Lin, Chin-Yew},
  booktitle={Text summarization branches out},
  pages={74--81},
  year={2004}
}

@inproceedings{caba2015activitynet,
  title={Activitynet: A large-scale video benchmark for human activity understanding},
  author={Caba Heilbron, Fabian and Escorcia, Victor and Ghanem, Bernard and Carlos Niebles, Juan},
  booktitle={Proceedings of the ieee conference on computer vision and pattern recognition},
  pages={961--970},
  year={2015}
}

@article{cheng2024spatialrgpt,
  title={Spatialrgpt: Grounded spatial reasoning in vision-language models},
  author={Cheng, An-Chieh and Yin, Hongxu and Fu, Yang and Guo, Qiushan and Yang, Ruihan and Kautz, Jan and Wang, Xiaolong and Liu, Sifei},
  journal={Advances in Neural Information Processing Systems},
  volume={37},
  pages={135062--135093},
  year={2024}
}

@article{fan2025video,
  title={Video-LLMs with Temporal Visual Screening},
  author={Fan, Zheyu and Liu, Jiateng and Zhang, Yuji and Wang, Zihan and Fung, Yi R and Li, Manling and Ji, Heng},
  journal={arXiv preprint arXiv:2508.21094},
  year={2025}
}

@article{xu2024slowfast,
  title={Slowfast-llava: A strong training-free baseline for video large language models},
  author={Xu, Mingze and Gao, Mingfei and Gan, Zhe and Chen, Hong-You and Lai, Zhengfeng and Gang, Haiming and Kang, Kai and Dehghan, Afshin},
  journal={arXiv preprint arXiv:2407.15841},
  year={2024}
}

@inproceedings{sun2025layoutvlm,
  title={Layoutvlm: Differentiable optimization of 3d layout via vision-language models},
  author={Sun, Fan-Yun and Liu, Weiyu and Gu, Siyi and Lim, Dylan and Bhat, Goutam and Tombari, Federico and Li, Manling and Haber, Nick and Wu, Jiajun},
  booktitle={Proceedings of the Computer Vision and Pattern Recognition Conference},
  pages={29469--29478},
  year={2025}
}

@inproceedings{bao2025exploiting,
  title={Exploiting vlm localizability and semantics for open vocabulary action detection},
  author={Bao, Wentao and Li, Kai and Chen, Yuxiao and Patel, Deep and Min, Martin Renqiang and Kong, Yu},
  booktitle={2025 IEEE/CVF Winter Conference on Applications of Computer Vision (WACV)},
  pages={8291--8301},
  year={2025},
  organization={IEEE}
}

@article{zhang2025morpheus,
  title={Morpheus: Benchmarking Physical Reasoning of Video Generative Models with Real Physical Experiments},
  author={Zhang, Chenyu and Cherniavskii, Daniil and Zadaianchuk, Andrii and Tragoudaras, Antonios and Vozikis, Antonios and Nijdam, Thijmen and Prinzhorn, Derck WE and Bodracska, Mark and Sebe, Nicu and Gavves, Efstratios},
  journal={arXiv preprint arXiv:2504.02918},
  year={2025}
}

@article{motamed2025travl,
  title={TRAVL: A Recipe for Making Video-Language Models Better Judges of Physics Implausibility},
  author={Motamed, Saman and Chen, Minghao and Van Gool, Luc and Laina, Iro},
  journal={arXiv preprint arXiv:2510.07550},
  year={2025}
}

@inproceedings{zhou2025vlm4d,
  title={Vlm4d: Towards spatiotemporal awareness in vision language models},
  author={Zhou, Shijie and Vilesov, Alexander and He, Xuehai and Wan, Ziyu and Zhang, Shuwang and Nagachandra, Aditya and Chang, Di and Chen, Dongdong and Wang, Xin Eric and Kadambi, Achuta},
  booktitle={Proceedings of the IEEE/CVF international conference on computer vision},
  pages={8600--8612},
  year={2025}
}

@inproceedings{fu2025video,
  title={Video-mme: The first-ever comprehensive evaluation benchmark of multi-modal llms in video analysis},
  author={Fu, Chaoyou and Dai, Yuhan and Luo, Yongdong and Li, Lei and Ren, Shuhuai and Zhang, Renrui and Wang, Zihan and Zhou, Chenyu and Shen, Yunhang and Zhang, Mengdan and others},
  booktitle={Proceedings of the Computer Vision and Pattern Recognition Conference},
  pages={24108--24118},
  year={2025}
}

@article{li2024wolf,
  title={Wolf: Dense Video Captioning with a World Summarization Framework},
  author={Li, Boyi and Zhu, Ligeng and Tian, Ran and Tan, Shuhan and Chen, Yuxiao and Lu, Yao and Cui, Yin and Veer, Sushant and Ehrlich, Max and Philion, Jonah and others},
  journal={arXiv preprint arXiv:2407.18908},
  year={2024}
}

@article{li2025eventvl,
  title={Eventvl: Understand event streams via multimodal large language model},
  author={Li, Pengteng and Lu, Yunfan and Song, Pinghao and Li, Wuyang and Yao, Huizai and Xiong, Hui},
  journal={arXiv preprint arXiv:2501.13707},
  year={2025}
}

@misc{cai2025holisticevaluationmultimodalllms,
      title={Holistic Evaluation of Multimodal LLMs on Spatial Intelligence}, 
      author={Zhongang Cai and Yubo Wang and Qingping Sun and Ruisi Wang and Chenyang Gu and Wanqi Yin and Zhiqian Lin and Zhitao Yang and Chen Wei and Oscar Qian and Hui En Pang and Xuanke Shi and Kewang Deng and Xiaoyang Han and Zukai Chen and Jiaqi Li and Xiangyu Fan and Hanming Deng and Lewei Lu and Bo Li and Ziwei Liu and Quan Wang and Dahua Lin and Lei Yang},
      year={2025},
      eprint={2508.13142},
      archivePrefix={arXiv},
      primaryClass={cs.CV},
      url={https://arxiv.org/abs/2508.13142}, 
}

@inproceedings{yang2025thinking,
  title={Thinking in space: How multimodal large language models see, remember, and recall spaces},
  author={Yang, Jihan and Yang, Shusheng and Gupta, Anjali W and Han, Rilyn and Fei-Fei, Li and Xie, Saining},
  booktitle={Proceedings of the Computer Vision and Pattern Recognition Conference},
  pages={10632--10643},
  year={2025}
}

@article{chen2024motionllm,
  title={Motionllm: Understanding human behaviors from human motions and videos},
  author={Chen, Ling-Hao and Lu, Shunlin and Zeng, Ailing and Zhang, Hao and Wang, Benyou and Zhang, Ruimao and Zhang, Lei},
  journal={arXiv preprint arXiv:2405.20340},
  year={2024}
}

@article{li2025survey,
  title={A survey of state of the art large vision language models: Alignment, benchmark, evaluations and challenges},
  author={Li, Zongxia and Wu, Xiyang and Du, Hongyang and Liu, Fuxiao and Nghiem, Huy and Shi, Guangyao},
  journal={arXiv preprint arXiv:2501.02189},
  year={2025}
}

@inproceedings{zhao2025mmvu,
  title={Mmvu: Measuring expert-level multi-discipline video understanding},
  author={Zhao, Yilun and Zhang, Haowei and Xie, Lujing and Hu, Tongyan and Gan, Guo and Long, Yitao and Hu, Zhiyuan and Chen, Weiyuan and Li, Chuhan and Xu, Zhijian and others},
  booktitle={Proceedings of the Computer Vision and Pattern Recognition Conference},
  pages={8475--8489},
  year={2025}
}

@inproceedings{li2024mvbench,
  title={Mvbench: A comprehensive multi-modal video understanding benchmark},
  author={Li, Kunchang and Wang, Yali and He, Yinan and Li, Yizhuo and Wang, Yi and Liu, Yi and Wang, Zun and Xu, Jilan and Chen, Guo and Luo, Ping and others},
  booktitle={Proceedings of the IEEE/CVF Conference on Computer Vision and Pattern Recognition},
  pages={22195--22206},
  year={2024}
}

@inproceedings{liang2025fine,
  title={Fine-grained spatiotemporal grounding on egocentric videos},
  author={Liang, Shuo and Zhong, Yiwu and Hu, Zi-Yuan and Tao, Yeyao and Wang, Liwei},
  booktitle={Proceedings of the IEEE/CVF International Conference on Computer Vision},
  pages={9385--9395},
  year={2025}
}

@inproceedings{balazadeh2025physics,
  title={Physics context builders: A modular framework for physical reasoning in vision-language models},
  author={Balazadeh, Vahid and Ataei, Mohammadmehdi and Cheong, Hyunmin and Khasahmadi, Amir Hosein and Krishnan, Rahul G},
  booktitle={Proceedings of the IEEE/CVF International Conference on Computer Vision},
  pages={7318--7328},
  year={2025}
}

@article{chi2025chimera,
  title={Chimera: Diagnosing Shortcut Learning in Visual-Language Understanding},
  author={Chi, Ziheng and Hou, Yifan and Pang, Chenxi and Cui, Shaobo and Akhtar, Mubashara and Sachan, Mrinmaya},
  journal={arXiv preprint arXiv:2509.22437},
  year={2025}
}

@inproceedings{li2025sti,
  title={Sti-bench: Are mllms ready for precise spatial-temporal world understanding?},
  author={Li, Yun and Zhang, Yiming and Lin, Tao and Liu, XiangRui and Cai, Wenxiao and Liu, Zheng and Zhao, Bo},
  booktitle={Proceedings of the IEEE/CVF International Conference on Computer Vision},
  pages={5622--5632},
  year={2025}
}

@article{wang2025spatialvid,
  title={Spatialvid: A large-scale video dataset with spatial annotations},
  author={Wang, Jiahao and Yuan, Yufeng and Zheng, Rujie and Lin, Youtian and Gao, Jian and Chen, Lin-Zhuo and Bao, Yajie and Zhang, Yi and Zeng, Chang and Zhou, Yanxi and others},
  journal={arXiv preprint arXiv:2509.09676},
  year={2025}
}
}

\clearpage
\setcounter{page}{1}
\maketitlesupplementary

\appendix




\section{Additional Details of \oursdata{}}
\label{sec:dataset_stats_supp}

This section supplements the dataset description in Section~\ref{sec:dataset} with expanded definitions and statistics.

\subsection{Dataset Composition and Metadata}

\oursdata{} aggregates three physics-centric video sources: $1{,}229$ samples ($14.7\%$) from MotionSight~\cite{du2025motionsight}, $4{,}009$ ($47.9\%$) from VideoPhy2~\cite{bansal2025videophy}, and $2{,}862$ ($34.2\%$) from VideoHallu~\cite{li2025videohallu}.
In total, the dataset comprises $6{,}093$ training and $2{,}268$ test examples drawn from $4{,}350$ unique videos. Videos average $545.8$ frames (${\sim}19.6$\,s at $27.4$ FPS) with a mean resolution of $1120{\times}702$ pixels. Due to overlap inherited from the source datasets, approximately $1.5\%$ of unique videos appear in both splits; however, no \mbox{(video, question)} pairs are shared across splits, so there is no direct QA leakage. All methods are evaluated on the same fixed test set with identical evaluation scripts. Importantly, \ours{} relies only on depth and tracking signals extracted from the input video at inference time and does not use any ground-truth test annotations.

\subsection{Question Categories}

We group all questions into five categories reflecting the type of physical reasoning required, ordered from basic perception to advanced comprehension. Table~\ref{tab:category_stats} shows the distribution.

\begin{table}[h]
\small
\centering
\begin{tabular}{lrr}
\toprule
\textbf{Category} & \textbf{Count} & \textbf{Pct.} \\
\midrule
SU (Spatial Understanding)            & 2,785 & 33.3\% \\
TU (Temporal Understanding)           & 1,633 & 19.5\% \\
PA (Physical Abnormality Detection)   & 1,432 & 17.1\% \\
PC (Physics Comprehension)            & 1,304 & 15.6\% \\
MAR (Motion \& Action Recognition)    & 1,205 & 14.4\% \\
\bottomrule
\end{tabular}
\vspace{-1mm}
\caption{\textbf{Distribution of question categories in \oursdata{}.}}
\label{tab:category_stats}
\end{table}

\noindent
\textit{(a) Spatial Understanding (SU).} Identifying objects and their geometric relationships, positions, and scene layouts.

\noindent
\textit{(b) Temporal Understanding (TU).} Interpreting how events evolve over time, including ordering, duration, and temporal dependencies.

\noindent
\textit{(c) Motion \& Action Recognition (MAR).} Detecting and characterizing object motions and agent actions across frames.

\noindent
\textit{(d) Physics Comprehension (PC).} Applying physical principles to infer, explain, or predict real-world dynamics.

\noindent
\textit{(e) Physical Abnormality Detection (PA).} Identifying motions or events that violate physical laws or exhibit implausible behavior.

\noindent
These categories are ordered from easy to challenging. VLMs must first establish spatial and temporal awareness, then recognize motion patterns, before progressing to physics comprehension and violation detection.

\subsection{Example and Question Types}

We categorize each video--question--answer pair along two complementary axes.

For \textbf{example type}, we distinguish \textit{positive} examples ($3{,}436$, $41.1\%$), where the video follows real-world physics, from \textit{negative} examples ($4{,}925$, $58.9\%$), where at least one part of the scene violates physical plausibility. Negative examples are more challenging because the model must rely on visual evidence rather than defaulting to language priors.

For \textbf{question type} (Figure~\ref{fig:dataset}), we separate \textit{factual} questions ($5{,}427$, $67.0\%$) from \textit{critical-thinking} questions ($2{,}673$, $33.0\%$). Factual questions provide explicit cues that focus attention on particular entities or events. Critical-thinking questions are less direct: they omit such cues and instead require the model to infer intent, identify salient events, or reason about the underlying physical dynamics.

\subsection{Motion Grounding Annotations}

Beyond the question, video, and ground-truth answer, each pair includes five layers of motion-grounding annotation: \textit{(1)} temporal video segmentation indexed by frame ranges; \textit{(2)} visual grounding for each queried entity, specified by entity name and bounding box; \textit{(3)} an entity-level temporal profile that tracks each grounded entity across the full video, persisting once detected and left blank in segments where the entity is absent; \textit{(4)} per-segment motion attributes, including the first and last observed 3D positions to capture coarse spatial layout; and \textit{(5)} 3D motion vectors representing each entity's temporal displacement. Together, these cues convert physics-intensive perceptual challenges into structured textual and mathematical representations that support more reliable physical reasoning.

\begin{figure*}
\begin{center}
\begin{tikzpicture}
\node [mybox,title=Prompt Template for Video-Language Models] (box){%
    \begin{minipage}{2\columnwidth}
        \raggedright


        \bluebold{Task Description:} \\
        The model receives a system instruction enforcing explicit reasoning and final
        answer formatting. Given a video, a question, and motion-grounding metadata,
        the model must produce detailed reasoning inside \texttt{<think>} tags and a 
        concise final answer inside \texttt{<answer>} tags.

        \vspace{4pt}
        \bluebold{Core Requirements:}
        \begin{itemize}[leftmargin=10pt]
            \item Use natural internal dialogue in the \texttt{<think>} section (e.g., ``let me think'', ``hmm'', ``wait'').
            \item Perform step-by-step reasoning validating spatial–temporal cues.
            \item Place the final answer \textbf{only} inside \texttt{<answer>} tags.
            \item Use \textbf{free-form answer format}: Provide a short text answer within the answer tags.
        \end{itemize}

        \par\noindent\rule{\textwidth}{0.8pt}

\bluebold{Motion-Grounding Information:} \\[2pt]
We use the following template to represent the motion-grounding information generated for VLMs. This template is used to fill the \texttt{\{motion\_grouding\_info\}} in the QA prompt template below: \\[4pt]

\begingroup
\ttfamily
\textbf{Entity \#1: <Entity Name>} \\[2pt]

{\small
* Segment \#1\: First Position <first\_position>, Motion Vector <motion>, Last Position <last\_position>, Bounding Box <bbox>, Frame <first\_frame>...<last\_frame> \\[2pt]

* Segment \#2\: First Position <first\_position>, Motion Vector <motion>, Last Position <last\_position>, Bounding Box <bbox>, Frame <first\_frame>...<last\_frame> \\[2pt]

\ldots
} \\[6pt]

\textbf{Entity \#2: <Entity Name>}
\ldots
\endgroup

\par\noindent\rule{\textwidth}{0.8pt}

\bluebold{Actual Prompt Used for Video QA:} \\[2pt]

\begingroup
\ttfamily
\textbf{Conversation Setup:} \\
A conversation between User and Assistant. The user asks a question, and the Assistant solves it. 
The Assistant first thinks about the reasoning process inside <think> </think> tags, 
then provides the final answer inside <answer> </answer> tags.\\[6pt]

\textbf{Question:} \\
<Question> \{question\} </Question>\\[6pt]

\textbf{Motion-Grounding Information:} \\
\{motion\_grouding\_info\}\\[6pt]

\textbf{Reasoning Instruction:} \\
Please think about this question as if you were a human pondering deeply. Use internal dialogue such as
``let me think'', ``wait'', ``hmm'', ``I see'', and include verification or self-reflection in the reasoning
process. Provide detailed reasoning in <think> </think>, then provide the final answer
in <answer> </answer>.\\[6pt]

\textbf{Free-form Answer Instruction:} \\
Please provide your text answer within the <answer> </answer> tags.
\endgroup

    \end{minipage}
};
\end{tikzpicture}%
\caption{\textbf{Prompt template used for motion-aware video question answering.}
The template first serializes entity-level motion grounding (positions, motion vectors, bounding boxes, and frame ranges) into text, then injects this context into a chain-of-thought style prompt that guides the VLM to reason in \texttt{<think>} tags and output its final prediction in standardized \texttt{<answer>} tags.}
\label{fig:vlm_prompt_template}
\end{center}
\end{figure*}

\begin{figure*}
\begin{center}
\begin{tikzpicture}
\node [mybox,title=Prompt Template for Evaluation] (box){%
    \begin{minipage}{2\columnwidth}
        \raggedright

        \bluebold{Task Description:} \\
        You are an intelligent teacher whose task is to evaluate the correctness of a model's answer
        to a question, given a reference ground-truth answer.

        \vspace{4pt}
        \bluebold{Inputs:}
        \begin{itemize}[leftmargin=10pt]
            \item \textbf{Question:} wrapped in \texttt{<Question> ... </Question>}
            \item \textbf{Ground-truth answer:} wrapped in \texttt{<GT> ... </GT>}
            \item \textbf{Model prediction:} wrapped in \texttt{<Answer> ... </Answer>}
        \end{itemize}

        \vspace{4pt}
        \bluebold{Evaluation Criteria:}
        \begin{itemize}[leftmargin=10pt]
            \item If the prediction \textbf{does not conflict} with the ground truth, output \texttt{<Eval> Correct </Eval>}.
            \item If the prediction \textbf{conflicts} with the ground truth, output \texttt{<Eval> Incorrect </Eval>}.
            \item If the correctness of the prediction is \textbf{unclear}, output \texttt{<Eval> Unclear </Eval>}.
            \item Reason carefully about the relationship between the prediction and the ground truth,
                  but keep the final evaluation \textbf{very brief}.
        \end{itemize}

        \vspace{4pt}
        \bluebold{Output Format:} \\
        Produce \emph{only} one of the following tokens as the final output: \\
        \hspace*{1em}\texttt{<Eval> Correct </Eval>} \\
        \hspace*{1em}\texttt{<Eval> Incorrect </Eval>} \\
        \hspace*{1em}\texttt{<Eval> Unclear </Eval>}

    \end{minipage}
};
\end{tikzpicture}%
\caption{\textbf{Prompt template used for automatic evaluation of model answers against ground-truth references.} The template presents the question, ground truth, and model output provided for LLM-as-a-judge evaluation and guides the evaluator to produce one of three outcomes, \textit{Correct}, \textit{Incorrect}, or \textit{Unclear}, ensuring reliable and consistent scoring across all predictions.}
\label{fig:evaluation_template}
\end{center}
\end{figure*}

\section{Implementation Details}

All experiments, including ablations, are conducted using full-parameter fine-tuning on 8 NVIDIA H100 (80GB) GPUs. Both Qwen2.5-VL-7B and LLaVA-OneVision-7B are trained under identical settings for fair comparison.
The GRPO post-training phase for each model requires approximately 9–12 hours, while the supervised fine-tuning (SFT) stage in our ablation studies completes within 2–3 hours.

\section{Visual Feature Representation Template}


Figure~\ref{fig:vlm_prompt_template} illustrates the prompt template used during both post-training and inference for VLMs in \oursdata{}. This template is designed to elicit structured reasoning for free-form video question answering. The motion-grounding information incorporated into the prompt is derived from the spatial–motion feature extraction module and the visual representation pipeline described in Section~\ref{sec:method}.

For each detected entity, denoted as \texttt{<Entity Name>}, we generate motion-grounding descriptors for every video segment in which the entity appears. These descriptors include the entity’s first and last 3D positions within the segment (\texttt{<first\_position>} and \texttt{<last\_position>}), the corresponding 3D motion vector (\texttt{<motion>}), the bounding box in the segment’s first frame (\texttt{<bbox>}), and the segment’s temporal extent indicated by the starting and ending frame indices (\texttt{<first\_frame>} and \texttt{<last\_frame>}).

\section{Evaluation Template}

We provide the LLM-as-a-judge evaluation template used in our experiments in Figure~\ref{fig:evaluation_template}. This template standardizes how we assess the correctness of model predictions by comparing each VLM’s answer against the ground-truth annotation and the corresponding question.

\section{Experiments on Real-world Video QA}
\label{sec:real_world_qa_supp}


As a supplement to the experiments in the main paper, we provide additional results on two real-world video question-answering benchmarks, MMVU~\cite{zhao2025mmvu} and MVBench~\cite{li2024mvbench}, to further assess the generalization ability of VLMs augmented with \ours{}. 
MMVU evaluates models on expert-level temporal, procedural, and interaction-centric reasoning, while MVBench is a widely used benchmark for temporal video understanding and action-centric tasks. 
Both datasets contain diverse video-based QA pairs captured from real-world scenarios.


Table~\ref{tab:real_world_data} reports the performance of VLMs integrated with \ours{} alongside state-of-the-art baselines. 
On MMVU, large-scale close-source models achieve substantially higher accuracy due to the benchmark’s emphasis on expert-level, interdisciplinary reasoning—highlighting the limitations of smaller 7B models. 
Qwen2.5-VL-7B+\ours{} improves over its baseline on both benchmarks and narrows the gap to close-source models, while LLaVA-OneVision-7B+\ours{} shows a performance drop on these out-of-distribution tasks. As also shown in Table~\ref{tab:mass_real_efficiency_merged_supp}, LLaVA is more sensitive to detection and tracking quality: when salient entities are missed or motion cues are noisy—common in MMVU and MVBench—performance can degrade, motivating more selective motion-profile extraction in future work.

\begin{table}[t]
\centering
\footnotesize
\resizebox{\columnwidth}{!}{
\begin{tabular}{l|cc}
\toprule
Model & MMVU (\%) & MVBench (\%) \\
\midrule
\multicolumn{3}{c}{\textit{Baselines}} \\
\midrule
\texttt{InternVL3.5-8B-Flash}      & 55.04 & 45.78 \\
\texttt{Qwen2.5-VL-7B}             & \textbf{64.00} & 51.30 \\
\texttt{LLaVA-OneVision-7B}        & 50.41 & \textbf{55.24} \\
\midrule
\multicolumn{3}{c}{\textit{Larger / Close-Source}} \\
\midrule
\texttt{GPT-4o}                    & \textbf{75.96} & \textbf{58.50} \\
\texttt{Gemini-2.5-Flash}          & 75.27 & 54.32 \\
\midrule
\multicolumn{3}{c}{\textit{VLMs with \ours{}}} \\
\midrule
\texttt{LLaVA-OneVision-7B + \ours{}} & 48.95 & 48.33 \\
\texttt{Qwen2.5-VL-7B + \ours{}}      & \textbf{65.76} & \textbf{61.59} \\
\bottomrule
\end{tabular}
}
\caption{\textbf{Overall accuracies (\%) on MMVU and MVBench.}
We report overall performance on two real-world video question-answering benchmarks, MMVU~\cite{zhao2025mmvu} and MVBench~\cite{li2024mvbench}. 
GPT-4o achieves the highest accuracy on both datasets. 
Qwen2.5-VL-7B+\ours{} improves over its baseline and narrows the gap to close-source models, while LLaVA-OneVision-7B+\ours{} is more sensitive to detection quality on these out-of-distribution tasks.}
\label{tab:real_world_data}
\vspace{-2mm}
\end{table}

\section{Noise-Induced Robustness}
\label{sec:noise_supp}

We inject two noise types into the motion profile to test spatial--motion grounding robustness: (1) \textbf{BBox/position noise}, perturbing bounding boxes and initial/final 3D positions (SAM-2 \& Grounding-DINO); (2) \textbf{Trajectory Noise}, perturbing 3D motion vectors (Co-Tracker). We apply a relatively large 10\% Gaussian perturbation to the motion vectors, positions, and bounding boxes, both separately and jointly. Table~\ref{tab:grpo_noise_ablation_supp} shows slight accuracy drops, yet \ours{}-augmented backbones still substantially outperform their baselines (Table~\ref{tab:input_variants_sft_grpo} in the main paper) across all noise settings, even after RFT. Because our physical comprehension is mostly qualitative and the motion profiles serve as supplementary cues rather than exact measurements, the model tolerates moderate value-level perturbations. Failures mainly occur when noise flips the sign of motion components, \ie, reversing positive/negative values, leading the model to make wrong assumptions about motion directions and further cascading into a misinterpretation of the entire physical process.

\begin{table}[t]
\centering
\footnotesize
\resizebox{\columnwidth}{!}{
\begin{tabular}{l|cc|ccc}
\toprule
 & \multicolumn{2}{c|}{\textbf{Baselines (\ours{})}} & \multicolumn{3}{c}{\textbf{Noise Injection}} \\
\multirow{-2}{*}{Model (+ \ours{})} & w/ & w/o & BBox & Motion & All \\
\midrule
\texttt{Qwen2.5-VL-7B} & 61.93 & 52.30 & 58.64 & 58.51 & 58.42 \\
\texttt{LLaVA-OneVision-7B}   & 59.24 & 45.79 & 54.50 & 54.19 & 54.19 \\
\bottomrule
\end{tabular}
}
\vspace{-10pt}
\caption{Noise-induced robustness evaluation on \ours{}-augmented models. BBox, Motion, and All columns report accuracy under 10\% Gaussian perturbations applied to bounding-box/position, 3D trajectory, and both components jointly. Baseline columns (\textit{w/} and \textit{w/o} \ours{}) are provided as reference.}
\label{tab:grpo_noise_ablation_supp}
\vspace{-12pt}
\end{table}

\section{Inference Time and Memory Usage}
\label{sec:efficiency_supp}

We report wall-clock inference time on 100 randomly sampled test cases on an NVIDIA L40S (Table~\ref{tab:mass_real_efficiency_merged_supp}). \ours{} increases per-query latency due to motion-profile extraction, while GPU memory increases only modestly. Our goal is to evaluate the benefit of explicit spatial--motion cues for physical understanding, not an optimized pipeline. Importantly, the motion profile is modular and cacheable per video, so it can be reused across multiple questions without re-extraction. Overhead can be further reduced via faster upstream modules and standard efficiency knobs (\eg, lower frame stride/resolution, smaller top-$K$ grounded entities), which we leave to future work.

\begin{table}[t]
\centering
\footnotesize
\resizebox{0.8\columnwidth}{!}{
\begin{tabular}{l|cc|cc}
\toprule
\multirow{2}{*}{Model} &
\multicolumn{2}{c|}{\textbf{Time (s/case)}} & \multicolumn{2}{c}{\textbf{GPU Mem (GB)}} \\
& w/o & w/ & w/o & w/ \\
\midrule
LLaVA-OV-7B      & 1.86 & 6.49  & 14.97 & 15.69 \\
Qwen2.5-VL-7B    & 5.83 & 10.46 & 15.45 & 16.18 \\
\bottomrule
\end{tabular}
}
\vspace{-8pt}
\caption{Per-query inference time (s) and GPU memory usage (GB) on an NVIDIA L40S over 100 randomly sampled test cases, with and without \ours{}.}
\label{tab:mass_real_efficiency_merged_supp}
\vspace{-15pt}
\end{table}

\section{Multi-object Tracking and Camera Motion}
\label{sec:tracking_camera_supp}

We agree that dense tracking in crowded scenes increases compute and may introduce failures. \ours{}-Bench is entity-centric: we track only grounded target entities (top-$K$, tunable; default 5), and our QA design and video filtering avoid highly crowded, many-entity cases. Camera ego-motion remains an open challenge; the current benchmark largely contains limited camera motion to focus on entity-centric spatial/motion grounding. \ours{} does not assume perfect tracks and degrades gracefully under noisy motion cues (as shown in the noise robustness analysis in Table~\ref{tab:grpo_noise_ablation_supp}). We plan to extend the benchmark and method to crowded multi-entity tracking and stronger camera motion handling in future work.


\section{Case Study}

\begin{figure*}[htbp]
\centering
\includegraphics[width=0.65\textwidth]{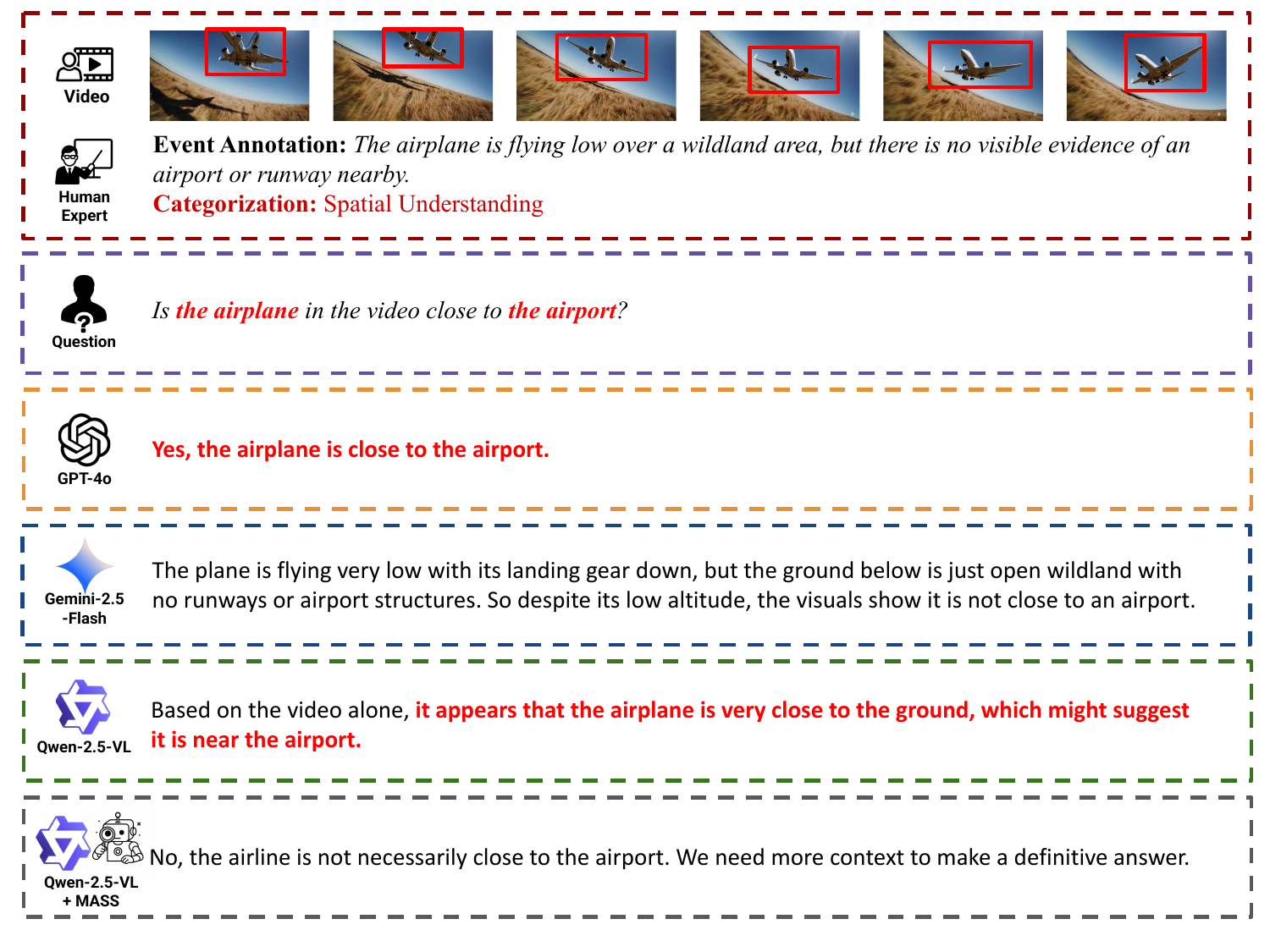}
\caption{\textbf{Video question-answering example from the Spatial Understanding (SU) category.}
We present physics reasoning and comprehension cases from state-of-the-art VLMs evaluated on spatial understanding tasks. Each example includes the video-generation prompt and human expert annotations, with visual grounding annotated (\textcolor{Red}{\textbf{Red}}), the corresponding questions (\textcolor{Purple}{\textbf{Purple}}), and model responses from GPT-4o (\textcolor{Orange}{\textbf{Orange}}), Gemini-2.5-Flash (\textcolor{Blue}{\textbf{Blue}}), Qwen2.5-VL (\textcolor{Green}{\textbf{Green}}), and Qwen2.5-VL~+~\ours{} (\textcolor{Gray}{\textbf{Gray}}). Hallucinated content and critical contextual errors are highlighted in \textcolor{Red}{\textbf{Red}}.}
\label{fig:showcase-SU}
\end{figure*}

\begin{figure*}[htbp]
\centering
\includegraphics[width=0.65\textwidth]{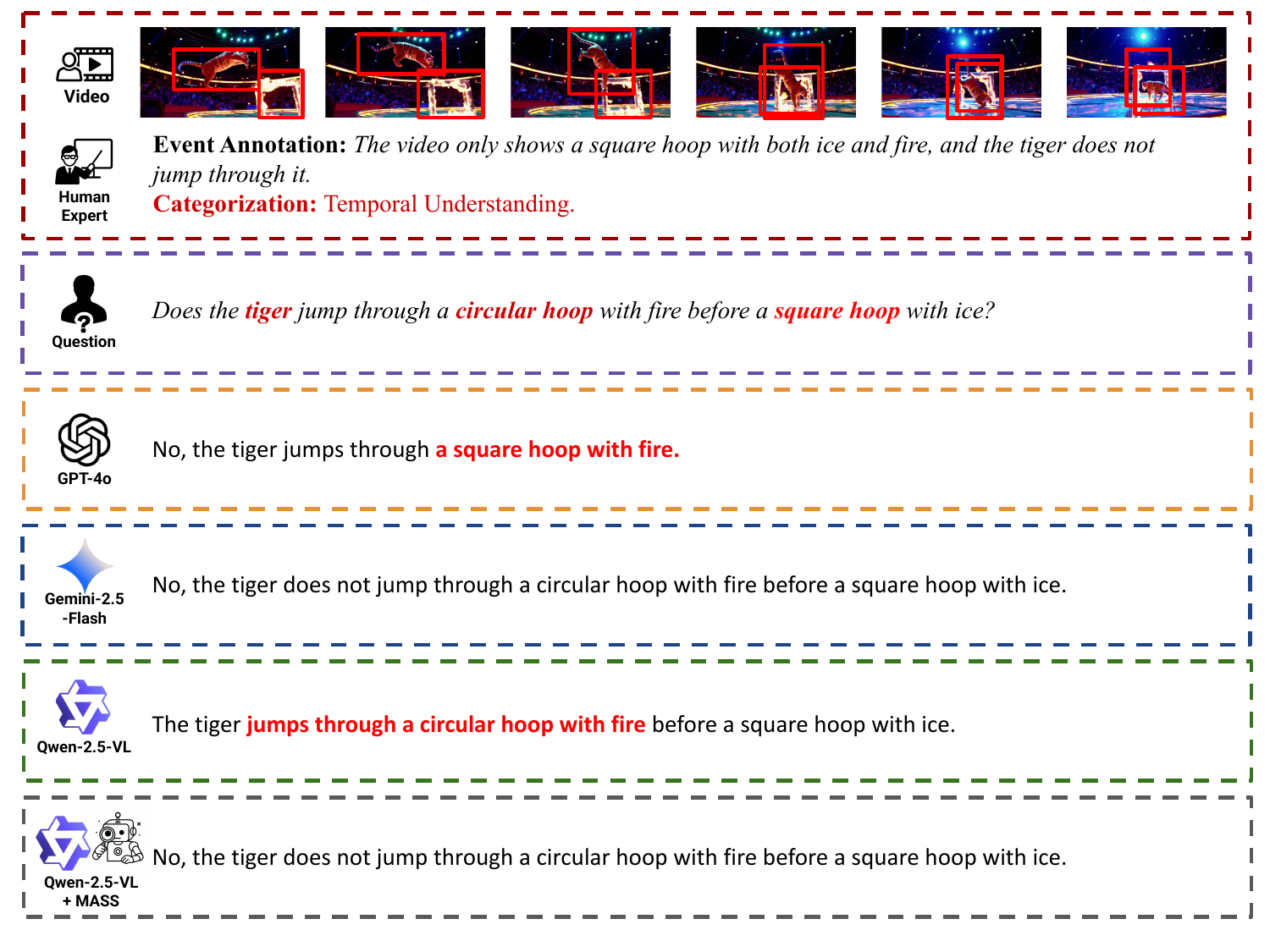}
\caption{\textbf{Video question-answering example from the Temporal Understanding (TU) category.}
We present physics reasoning and comprehension cases from state-of-the-art VLMs evaluated on temporal understanding tasks. Each example includes the video-generation prompt and human expert annotations, with visual grounding annotated (\textcolor{Red}{\textbf{Red}}), the corresponding questions (\textcolor{Purple}{\textbf{Purple}}), and model responses from GPT-4o (\textcolor{Orange}{\textbf{Orange}}), Gemini-2.5-Flash (\textcolor{Blue}{\textbf{Blue}}), Qwen2.5-VL (\textcolor{Green}{\textbf{Green}}), and Qwen2.5-VL~+~\ours{} (\textcolor{Gray}{\textbf{Gray}}). Hallucinated content and critical contextual errors are highlighted in \textcolor{Red}{\textbf{Red}}.}
\label{fig:showcase-TU}
\end{figure*}

\begin{figure*}[htbp]
\centering
\includegraphics[width=0.65\textwidth]{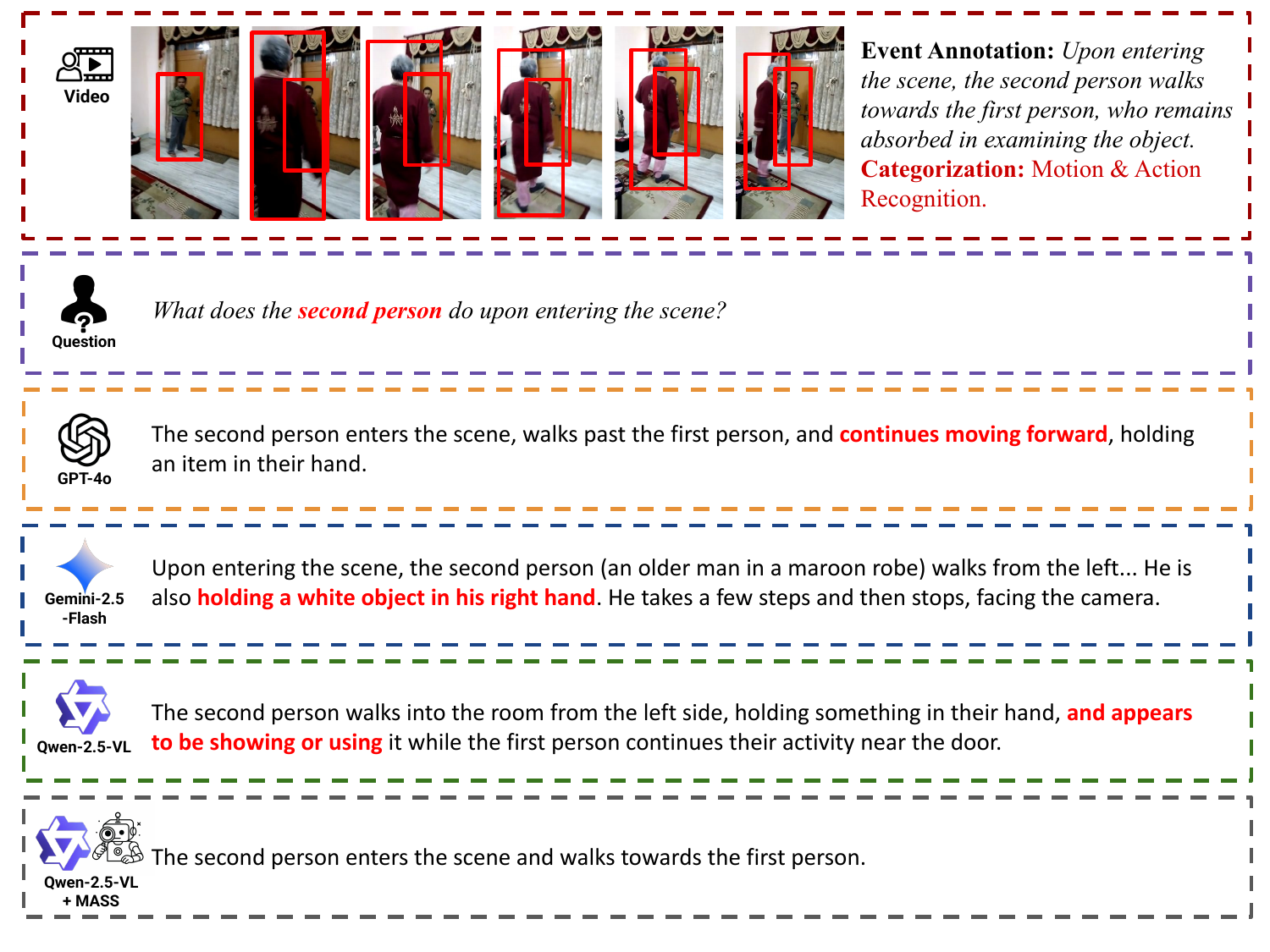}
\caption{\textbf{Video question-answering example from the Motion and Action Recognition (MAR) category.}
We present physics reasoning and comprehension cases from state-of-the-art VLMs evaluated on motion and action recognition tasks. Each example includes the video-generation prompt and human expert annotations, with visual grounding annotated (\textcolor{Red}{\textbf{Red}}), the corresponding questions (\textcolor{Purple}{\textbf{Purple}}), and model responses from GPT-4o (\textcolor{Orange}{\textbf{Orange}}), Gemini-2.5-Flash (\textcolor{Blue}{\textbf{Blue}}), Qwen2.5-VL (\textcolor{Green}{\textbf{Green}}), and Qwen2.5-VL~+~\ours{} (\textcolor{Gray}{\textbf{Gray}}). Hallucinated content and critical contextual errors are highlighted in \textcolor{Red}{\textbf{Red}}.}
\label{fig:showcase-mar}
\end{figure*}

\begin{figure*}[htbp]
\centering
\includegraphics[width=0.65\textwidth]{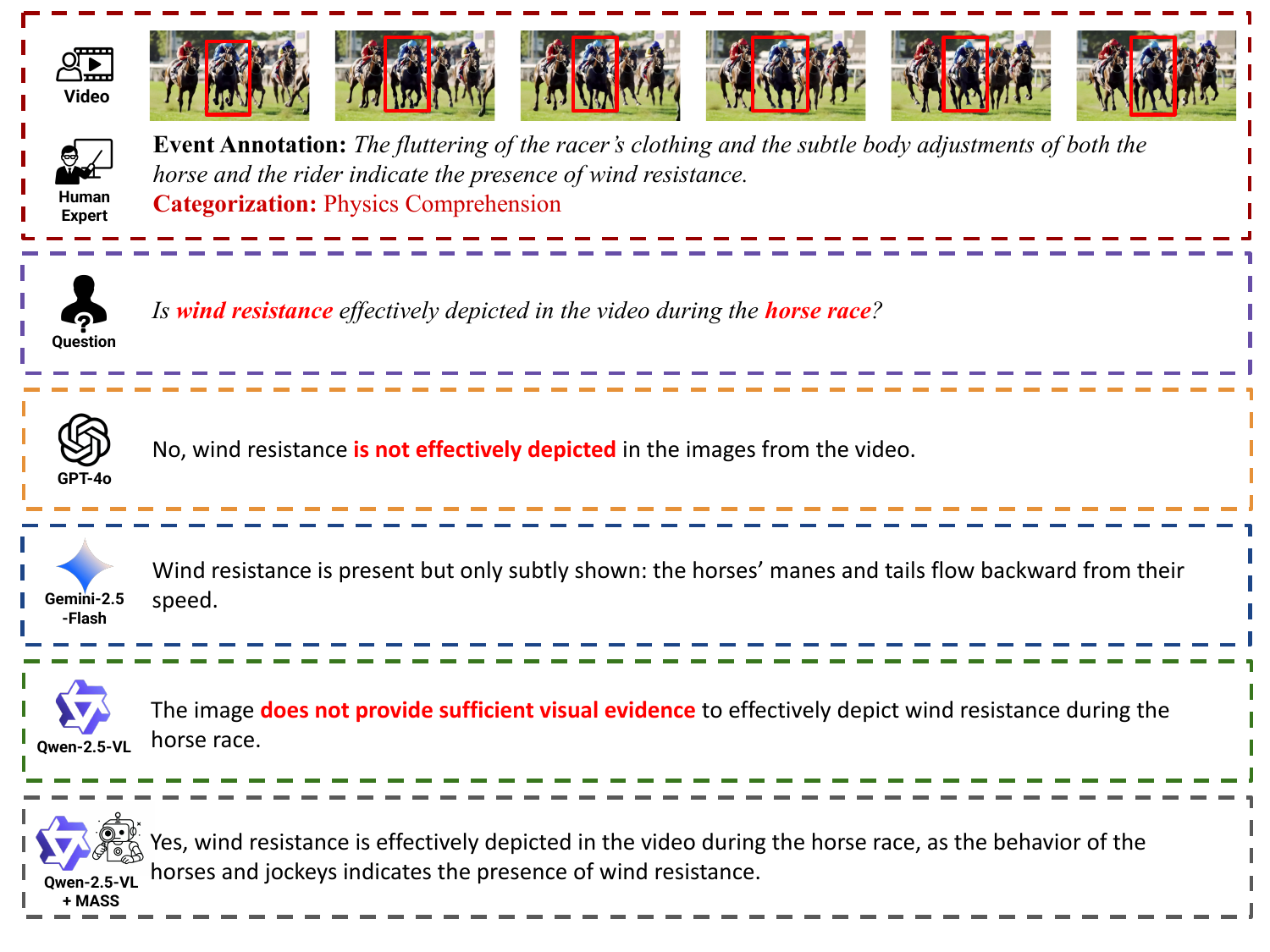}
\caption{\textbf{Video question-answering example from the Physics Comprehension (PC) category.}
We present physics reasoning and comprehension cases from state-of-the-art VLMs evaluated on physics comprehension tasks. Each example includes the video-generation prompt and human expert annotations, with visual grounding annotated (\textcolor{Red}{\textbf{Red}}), the corresponding questions (\textcolor{Purple}{\textbf{Purple}}), and model responses from GPT-4o (\textcolor{Orange}{\textbf{Orange}}), Gemini-2.5-Flash (\textcolor{Blue}{\textbf{Blue}}), Qwen2.5-VL (\textcolor{Green}{\textbf{Green}}), and Qwen2.5-VL~+~\ours{} (\textcolor{Gray}{\textbf{Gray}}). Hallucinated content and critical contextual errors are highlighted in \textcolor{Red}{\textbf{Red}}.}
\label{fig:showcase-pc}
\end{figure*}

\begin{figure*}[htbp]
\centering
\includegraphics[width=0.65\textwidth]{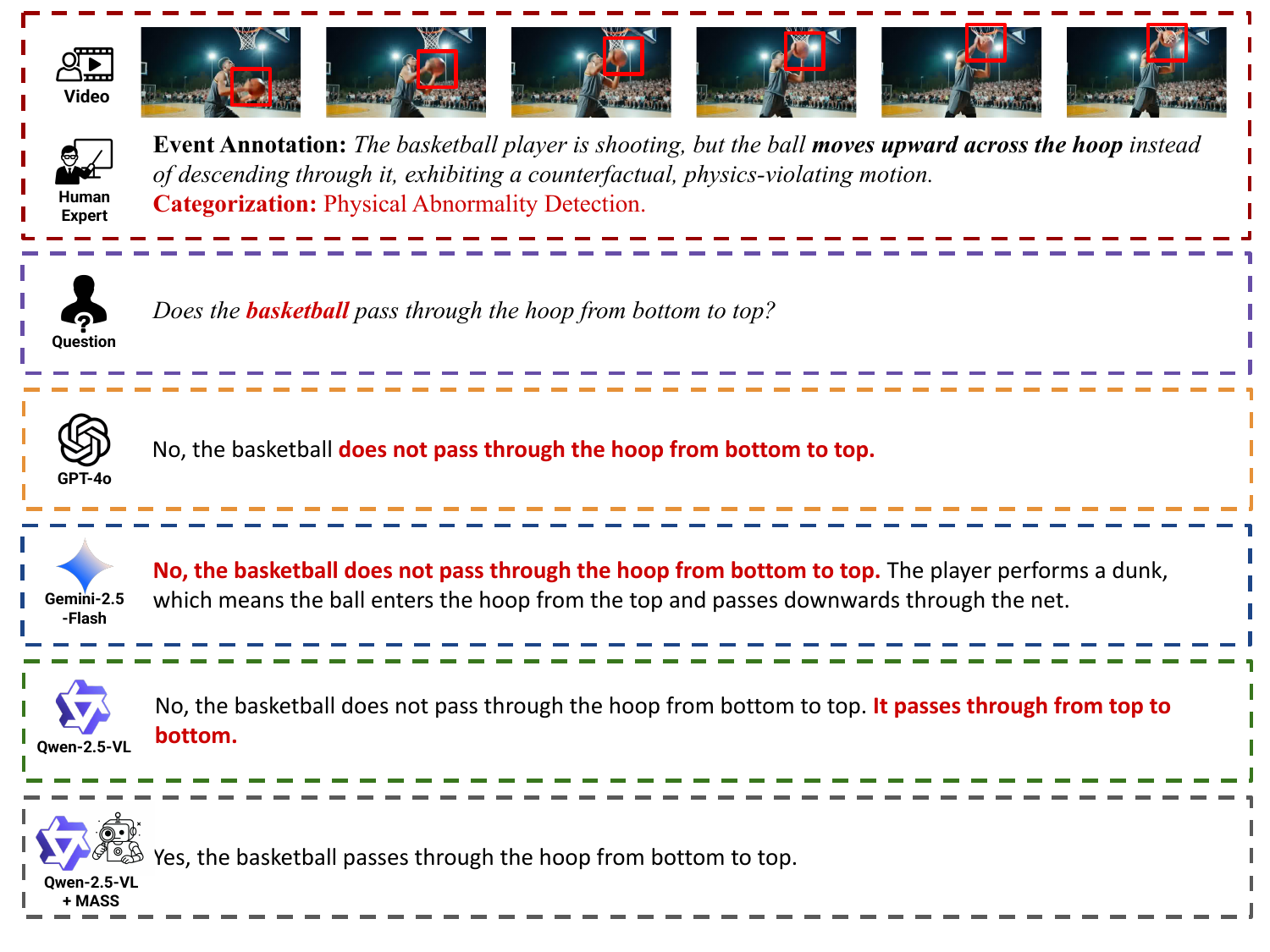}
\caption{\textbf{Video question-answering example from the Physics Abnormality Detection (PA) category.}
We present physics reasoning and comprehension cases from state-of-the-art VLMs evaluated on physics abnormality detection tasks. Each example includes the video-generation prompt and human expert annotations, with visual grounding annotated (\textcolor{Red}{\textbf{Red}}), the corresponding questions (\textcolor{Purple}{\textbf{Purple}}), and model responses from GPT-4o (\textcolor{Orange}{\textbf{Orange}}), Gemini-2.5-Flash (\textcolor{Blue}{\textbf{Blue}}), Qwen2.5-VL (\textcolor{Green}{\textbf{Green}}), and Qwen2.5-VL~+~\ours{} (\textcolor{Gray}{\textbf{Gray}}). Hallucinated content and critical contextual errors are highlighted in \textcolor{Red}{\textbf{Red}}.}

\label{fig:showcase-PA}
\end{figure*}

In this section, we present additional qualitative examples of video question answering produced by several SoRA VLMs across the categories defined in \oursdata{}, as shown in Figures~\ref{fig:showcase-SU}--\ref{fig:showcase-PA}. Hallucinated or incorrect predictions are highlighted in red. Each example is accompanied by expert human annotations describing the ground-truth physics-driven dynamics, providing a clear reference for evaluating model behavior and identifying failure modes.

\newpage

\end{document}